\documentclass[]{fairmeta}

\usepackage{titletoc}
\usepackage{hyperref}
\usepackage{url}
\usepackage{graphicx}
\usepackage{amsmath}
\usepackage{algorithm}
\usepackage{algpseudocode}
\usepackage{booktabs}
\usepackage{wrapfig}
\usepackage{subcaption}
\usepackage{xspace}
\usepackage{cleveref}
\usepackage{xcolor}
\usepackage{listings}
\usepackage{lscape}
\usepackage[font=small]{caption}
\renewcommand{\paragraph}[1]{\vspace{.1em}\noindent\textbf{#1}}

\newcommand{\ours}{\textsc{ScaleMAS}\xspace}

\title{Scaling Multi-Agent Intelligence with Substrate-Centric Coordination}

\author[1,3,*]{Zhenting Qi}
\author[2]{Zhuokai Zhao}
\author[1]{Rulin Shao}
\author[1]{Kerem Oktar}
\author[1]{Zhiyuan Yao}
\author[3]{Himabindu Lakkaraju}
\author[1]{Wen-tau Yih}
\author[1,\dagger]{Ansong Ni}
\author[1,\dagger]{Bhargavi Paranjape}
\author[3,4,\dagger]{Yilun Du}

\affiliation[1]{FAIR at Meta}
\affiliation[2]{Meta AI}
\affiliation[3]{Harvard}
\affiliation[4]{Kempner Institute}

\contribution[*]{Work done at Meta}
\contribution[\dagger]{Joint last author}

\abstract{Existing large language model (LLM) based multi-agent systems have been shown to scale poorly, and we identify three problems: reading burden that grows with the group, useful information diluted by shared work of unknown quality, and fragility to failed agents. We argue that the problems are largely due to agent-centric designs: the coordination structure is wired into fixed workflows, central coordinators, or peer-to-peer messaging, and recent blackboard-style methods that move the coordination into a shared medium relieve only the last problem. To close the gaps, we build \ours, where asynchronous agents interact with a shared datastore of linked messages via three operations: publishing messages with optional references to earlier ones that they build on, searching for messages by content, and visiting messages by address. Search returns at most $k$ messages, so an agent's read stays bounded at any group size; results are ranked by relevance and by how often they are cited, which we find to be an effective proxy for quality; and messages persist after their producers leave, so the group carries on when individual agents fail. 
Across four agentic tasks, \ours outperforms agent-centric and blackboard-style baselines at lower inference compute, continues to scale up to 128 agents, and tolerates losing agents midway.}

\date{\today}
\correspondence{Zhenting Qi at \email{zhentingqi@g.harvard.edu}}

\begin{document}

\maketitle

\begin{figure}[!htb]
    \centering
    \includegraphics[width=\linewidth]{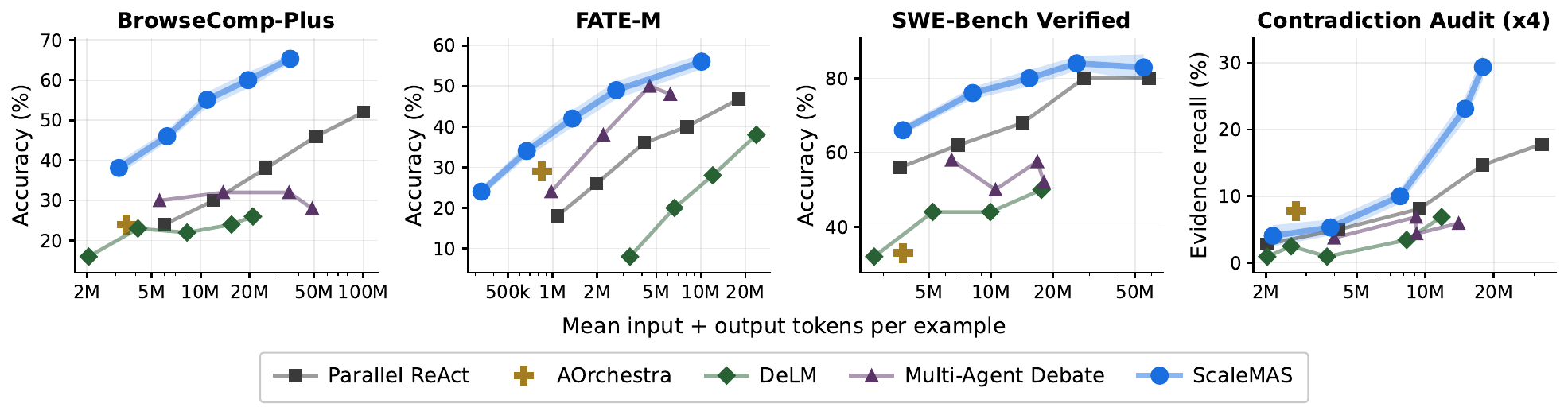}
    \caption{\textbf{\ours shows stronger scalability.} Task performance v.s. inference compute per example (log scale) with Qwen3.5-9B. Each point is one group size, scaled from 1 to 2, 4, 8, and 16 agents; AOrchestra, whose coordinator decides how many sub-agents to launch, is a single point. At matched compute, \ours reaches higher performance than Multi-Agent Debate (all-to-all message-passing), AOrchestra (coordinator-based), DeLM (blackboard-style shared context), and parallel ReAct agents without communication, and it improves with group size up to eight agents on every benchmark.} 
    \label{fig:performance}
    \vspace{-10pt}
\end{figure}

\section{Introduction}
\label{sec:intro}

Scaling a group of cooperative agents should increase its collective capability, but this benefit is not automatic in current large language model (LLM) based multi-agent systems (MAS). Empirical studies report superlinearly increased inference compute, diminishing or negative returns from additional agents, and settings in which a group underperforms a single agent given the same inference budget~\citep{smit2023should, silobench2026, kim2026capable, zheng2026thought}. 

We identify three major problems that grow with the number of agents and that many of the current designs leave unsolved. First, what each agent must read increases with the group: message-passing systems deliver all senders' outputs to a receiver agent~\citep{du2024debate, liu2026message}, and coordinator-based systems route every sub-agent's result into a central agent's context~\citep{su2025toolorchestra, ruan2026aorchestra}, thus a larger group floods its readers with input that could exceed context limits. Second, the useful information dilutes as the group grows: even with bounded reading, an agent must find the few relevant results in an ever larger body of partial work shared by others, and it is hard to tell the quality of the shared information. Third, failure becomes the routine rather than the exception: the more agents run, the more likely some of them stall or crash, yet workflow-based systems fix each agent's role and successor at design time~\citep{hong2024metagpt, qian2024chatdev} and coordinator-based systems route everything through one agent~\citep{cemri2026multi}, so a lost agent could easily cause the system to crash.

\begin{figure}[t]
\centering
    \includegraphics[width=\linewidth]{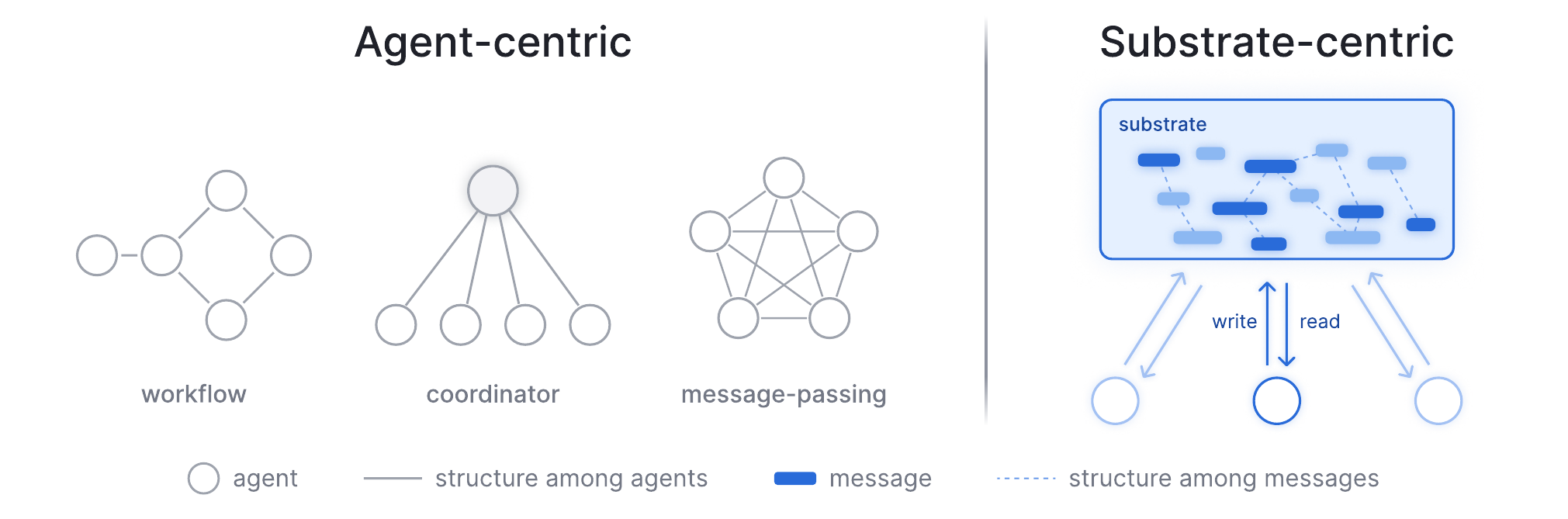}
    \caption{\textbf{Two design principles for multi-agent coordination.} \textbf{Left:} agent-centric designs wire coordination structure into the agents. A workflow-based system fixes agents' successors, a coordinator-based system routes results through one main agent, and a message-passing system lets agents address others. \textbf{Right:} substrate-centric designs place the coordination in a shared medium. Messages are pushed to a datastore, organized with certain structures; agents interact only with the medium by writing to and reading from it.}   
    \label{fig:teaser}
    \vspace{-15pt}
\end{figure}

What these methods share is that they adopt \textit{agent-centric} designs by wiring coordination structures into agents' instructions and loops. 
Another design principle has been adopted for decades since early blackboard systems~\citep{newell1962problems, niibbpart1, erman1980hearsay}: rather than engineering agents for better coordination, one can design the \textit{substrate}\footnote{Throughout this paper, we define an agent substrate as a combination of: a datastore that organizes agents' messages, operations for reads and writes, and deterministic rules that connect the datastore and the operations.} where agents only interact with a shared environment and self-coordinate (\Cref{fig:teaser}).
Recent work in LLM-based multi-agent systems has been trying to adopt this principle by giving LLM agents either a shared context~\citep{mao2026shared, park2026scalingdiscoverytesttimecommunication, qi2026coact} or a shared filesystem~\citep{qu2026coral, kim2026multi}.
This could alleviate the third problem: no agent waits on a particular peer, so a stalled agent blocks no one; and results persist after their producers leave, so a failed agent loses only its unfinished work.
However, it leaves the first two problems unsolved. A shared context grows linearly with the group size, which still could cause context overflow; a shared filesystem bounds nothing, and leaves each agent to search a growing pile of files with no signal of their quality.
A scalable substrate for LLM agents therefore needs more than a shared medium: it must bound each agent's reading as the group's output grows, while maintaining effective information discovery for every agent.

Both requirements have a working precedent at the scale of billions of readers and writers: the Web. Its scalability comes in part from avoiding readers having to read everything: search bounds attention by returning a small set of relevant results, while hypertextual links make information navigable on demand and also provide signals of its importance~\citep{brin1998anatomy, Page1999ThePC}. Drawing on these principles, we build \ours, a task-agnostic substrate on which LLM agents self-coordinate at scale. The agent substrate maintains a datastore of persistent messages interconnected through references, and a search engine indexed over them. Agents run asynchronously, and each interacts only with the substrate through three operations: it \textit{publishes} messages to the datastore, optionally referencing earlier ones; it \textit{searches} for messages by content and gets at most $k$ returned results; and it \textit{visits} messages for details by address. When a new message references earlier ones, the referenced messages gain importance weight, which combines with lexical relevance to rank subsequent search results. Every message is displayed with its references and with the later messages that reference it, but never with its author or its weight.

Several properties follow from the design. First, an agent takes in bounded messages per read operation, regardless of how many agents are writing or how large the datastore has grown. Second, the importance of shared information is earned through reuse: weight accumulates only when others build on a message, so search surfaces what the group has actually relied on, and the signal is enhanced as the datastore grows. Third, membership is flexible to change. Because messages outlive their producers, a newcomer reads the datastore and can catch up by searching for and visiting messages, and an agent that leaves takes nothing with it. Moreover, lineage of shared results is inspectable, as each message carries references and its citers, and a reader can check what a claim rests on before building on it. 

We evaluate \ours on four agentic tasks spanning deep research~\citep{chen2025browsecomp}, theorem proving~\citep{jiang2026fate}, software engineering~\citep{jimenez2024swe}, and contradiction audit, a benchmark we construct in which agents identify contradictions in large document corpora. We use two LLMs as backbones and compare against multi-agent debate (all-to-all message-passing)~\citep{du2024debate}, DeLM (blackboard-style shared context)~\citep{mao2026shared}, AOrchestra (coordinator-based delegation)~\citep{ruan2026aorchestra}, and parallel ReAct agents without communication~\citep{yao2023react}. Our experiments examine how collective performance and inference compute change with group size, with additional scaling studies extending to $128$ agents. We also evaluate resilience to unreliable membership and ablate citation-based ranking to assess its contribution. A group of $N$ agents with one failed agent performs comparably to $N-1$ healthy agents, indicating that the remaining agents can continue cooperating despite the loss of individual participants. These results suggest that multi-agent systems scale further when coordination is designed into the substrate the agents share rather than wired into the agents themselves.

\section{Method}
\label{sec:method}
\suppressfloats[t]

\begin{figure}[t]
\centering
\includegraphics[width=\linewidth]{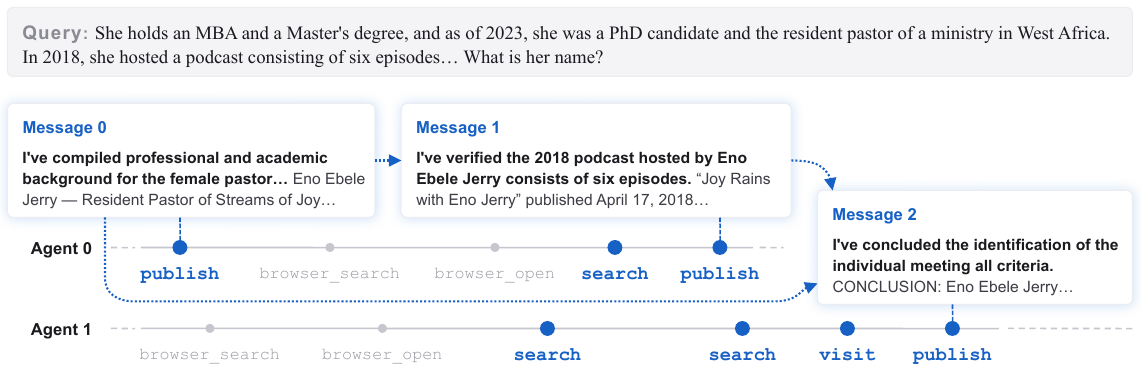}
\caption{\textbf{A minimal example of two agents working through the substrate} on a deep research query. Each row is one agent's trajectory, with the three operations in blue and the task's own tools in gray. A dashed stem marks the step that published a message, and a dotted arrow points from a referenced message to the message that cites it.}
\label{fig:pages}
\vspace{-15pt}
\end{figure}

\ours is an agent substrate with three parts: 1) a datastore of messages, 2) three operations that agents use to write to it and read from it, and 3) deterministic rules that decide what is written and what a read returns. $N$ agents run the same harness under the same prompt. Each is an ordinary tool-using LLM agent that keeps the tools of its task and gains the three extra operations, which are its only access to the datastore. 
No agent holds a role, sends messages to another agent, or assigns work to other agents, and the agents run asynchronously. 
\Cref{fig:pages} shows an example that traces two agents through the substrate.

\subsection{The datastore}

The datastore $D$ is an append-only sequence of messages. A message $m=(u,a,r,R)$ records one piece of completed work: the sub-task $u$ the agent carried out, stated in the words someone about to do the same work would search for; the artifact $a$ that the work produced; a short rationale $r$; and a set $R$ of references to what the work built on. A reference names either an earlier message or a resource exposed by the task's own tools, such as a corpus snippet or a repository file. Each message receives the next address when it is written and is never edited or deleted, so a reference always resolves to the text it was written against, and the references form a directed acyclic graph over $D$ that grows only at its newest end. Each message also carries a citation weight $s_m\in[0,1)$, which the rules maintain and which search uses for ranking. The sub-task and artifact together are limited to $256$ or $1{,}024$ tokens depending on the benchmark (\Cref{tab:hyper}), and a message carries at most $32$ references.

\subsection{The operations}

\texttt{publish}$(u,a,r,R)$ appends a message and makes it readable by every agent from that moment on. The two read operations match the two ways a reader can know what it wants. \texttt{search}$(q)$ takes a free-text description of what the agent needs and returns the $k$ highest-ranked messages. \texttt{visit}$(m_1,\dots,m_j)$ returns the named messages, for an agent that a search result or a message's references has already pointed somewhere specific. Both return each message as its four fields, the messages it references, and the most recent of the messages that reference it, up to eight. The author and the citation weight are never shown, so a reader judges a message by its content and its links. An agent ends by submitting its answer together with references and a rationale in the same form as a message, except on SWE-bench Verified, where the answer is a patch. The messages an answer cites, followed transitively through their references, form the answer's \emph{evidence chain}, which Section~\ref{sec:analysis} uses to trace where correct answers come from.

\paragraph{A walk-through.}
\Cref{fig:pages} shows how the operations combine on a deep research query. Agent~0
publishes Message~0, a candidate answer with the background it has gathered, then
browses further, searches the datastore, and publishes Message~1, which verifies one
clue of the query and cites Message~0. Agent~1 has meanwhile browsed on its own; it
searches the datastore, finds Message~1, follows its reference to Message~0 with
\texttt{visit}, and publishes Message~2, which combines the two into a conclusion and
cites both. Neither agent addresses the other or waits for it: Agent~1 learns of
Agent~0's work only by reading the datastore, and its final message rests on evidence
it did not gather itself.

\subsection{The rules}
\label{sec:ranking}

\paragraph{What is written.}
\texttt{publish} accepts a message only if every reference names something that exists and the rationale names every reference and states what it contributed. A message that breaks either rule is rejected with the offending reference named, and nothing is written. The first rule keeps references pointing backward in time. The second turns a reference into a stated dependency that a reader can check against the cited text, rather than an incidental mention. These are the only constraints \ours places on what an agent writes.

\paragraph{How weight is assigned.}
A new message starts at weight $s_0=0$. When a message references $m$, the weight of $m$ moves a fraction $\alpha$ of the way toward one, $s_m \leftarrow s_m+\alpha(1-s_m)$, once per citing message, and at every step of the group all weights decay, $s_m\leftarrow\gamma s_m$. A message that later work builds on therefore gains weight, and one that the group stops building on loses it. References to task resources carry no weight.

\paragraph{What a read returns.}
The search index ranks messages by
\begin{equation}
\mathrm{score}(q,m) = s_m + \frac{\mathrm{Rel}(q,m)}{\max_{m'\in D}\mathrm{Rel}(q,m')}.
\label{eq:score}
\end{equation}
The citation weight plus a retrieval score normalized to $[0,1]$; any retrieval
function can serve as $\mathrm{Rel}$, and we use BM25 \citep{robertson2009probabilistic}. Only messages with a positive
retrieval score are ranked, ties go to the earlier message, and the top $k$ are
returned, so the weight reorders relevant messages but cannot surface an unrelated one. The citation weight follows the idea of PageRank~\citep{brin1998anatomy, Page1999ThePC}, with links written by the agents as they work. \texttt{visit} likewise accepts at most $k$ addresses per call. Every read therefore adds at most $k$ messages of bounded length to an agent's context, however many agents are writing and however large $D$ has grown: an agent's intake grows with how often it reads, which the agent decides, and not with the size of the group.

\section{Experiments}

\subsection{Contradiction Audit}
\label{sec:contradiction-audit}

Few existing tasks allow multi-agent systems to be studied at different scales under control. Also, on many existing agentic benchmarks a single agent already performs well, which leaves little room to measure how performance changes with group size. We therefore construct a new task, Contradiction Audit, which is designed to reward combining information spread across agents: The agents receive a large software documentation corpus with injected factual contradictions, and are tasked with reporting as many pairs of contradicting passages as they can find. The corpus can grow while the injected contradictions stay fixed, so difficulty can be varied without changing the task. More details are given in \Cref{app:ca}.

\paragraph{Benchmark construction.}
We source the Markdown and reStructuredText documentation files from 16 open-source software projects, including CPython, the Linux kernel, and PyTorch, and exclude files that disagree with the rest by design (changelogs, release notes, deprecation notices). We use GPT-5.6 to select pairs of document snippets on a common subject and rewrite the prose of one to contradict the other.
An edit is kept only if a \texttt{diff} against the original snippet shows that it changes a few sentences and leaves the rest intact, and a second round of LLM judge (also with GPT-5.6) confirms that the edited snippet reads naturally, is not trivial (so that LLMs will not find it by memorization), is self-consistent, and truly conflicts with
its partner. The LLM judge will run four times, and a snippet will be accepted iff. all four judges are positive. Each corpus receives twenty injected contradictions over forty distinct snippets, ten within one file and ten across files.

\paragraph{Difficulty and scoring.}
Level \texttt{x}$n$ pads the forty contradiction snippets with seeded random
distractors to $40n$ snippets, so the corpus grows while the answers stay fixed.
For example, a \texttt{x}$8$ split means besides the 20 pairs of snippets that a system should really focus on, there are $40\times8-40=280$ other irrelevant snippets as distractions.
We use a relaxed metric for evaluation: A contradiction counts as found when a reported pair includes its edited snippet, and extra reports are not penalized. The score is the fraction of the twenty found,
averaged over 16 corpora. In the example of Table~\ref{tab:ca-example}, each passage reads naturally on its own and the two sit in different documents, so detecting the contradiction requires reading one snippet and locating the other.

\begin{table}[t]
\centering
\small
\newcommand{\injected}[1]{\textcolor[rgb]{0.55,0.15,0.15}{#1}}
\caption{\textbf{Example of a cross-file contradiction from the SymPy documentation corpus.} The injected text is in red, and the clashing claims are underlined.}
\label{tab:ca-example}
\begin{tabular}{@{}p{0.475\linewidth}@{\hspace{0.05\linewidth}}p{0.475\linewidth}@{}}
\toprule
\texttt{modules/solvers/solveset.rst} & \texttt{modules/sets.rst} (edited) \\
\midrule
\raggedright \ldots\ Set objects conform to Pythonic conventions when possible,
i.e., \texttt{x in A} and \texttt{for i in A} \underline{both work when they can
be computed}.
&
\raggedright \ldots\injected{By contrast, \texttt{x in y}
\underline{always gives \texttt{TypeError}} for SymPy sets, even when the membership
could easily be computed.} \tabularnewline
\bottomrule
\end{tabular}
\vspace{-10pt}
\end{table}

\subsection{Setup}
\label{sec:setup}

\paragraph{Benchmarks.}
Beyond Contradiction Audit, we select three more benchmarks that stress a group of LLM agents in different ways.
\textbf{BrowseComp-Plus} \citep{chen2025browsecomp} is search-intensive: the evidence
for each question is scattered across a large corpus of web documents,
reached through a search index. \textbf{FATE}
\citep{jiang2026fate} is reasoning-intensive: each theorem proving problem asks for a complete Lean~4 proof that the compiler checks.
\textbf{SWE-bench Verified} \citep{jimenez2024swe} measures the resolution of real GitHub issues, and requires multiple steps of verification and revision.
\ours is harness-agnostic: we run it on a standard ReAct loop~\citep{yao2023react}, implemented with the Hugging Face smolagents library\footnote{\url{https://github.com/huggingface/smolagents}}, for BrowseComp-Plus, FATE, and Contradiction Audit, and on mini-SWE-agent \citep{yang2024swe} for SWE-bench Verified. More benchmark details are presented in \Cref{app:eval}.

\paragraph{Baselines and parity.}
We compare against the following methods:
Parallel \textbf{ReAct} \citep{yao2023react} runs $N$ agents without communication between them.
\textbf{Multi-Agent Debate (MAD)} \citep{du2024debate} is agent-centric with all-to-all message passing: agents exchange their
intermediate work, each reading what every other produced. \textbf{AOrchestra}
\citep{ruan2026aorchestra} is agent-centric with a central coordinator: a main agent decomposes the task and
delegates to sub-agents, each of which reports back once. \textbf{DeLM}
\citep{mao2026shared} is blackboard-style: agents write to a shared context that all of them read, without addressing one another.
For parallel ReAct, MAD, DeLM, and \ours, an agent stops once it outputs a final answer, and the system stops when all agents stop. For AOrchestra, the system stops once the coordinator agent stops.
Every method runs with \textbf{Qwen3.5-9B} on all four benchmarks and with \textbf{Claude-5-Sonnet} on all but SWE-bench Verified. The stronger model gets the harder benchmark splits (FATE-H and Contradiction Audit \texttt{x8}; Qwen3.5-9B gets FATE-M and \texttt{x4}). 
All methods share identical benchmark task prompts, per-benchmark limit on action steps, and the benchmark's tools and evaluation harness.

\subsection{Main results}

\begin{figure}[t]
    \centering
    \newlength{\figrowh}\setlength{\figrowh}{0.238\textwidth}
    \begin{minipage}[t]{0.675\linewidth}
        \centering
        \includegraphics[height=\figrowh]{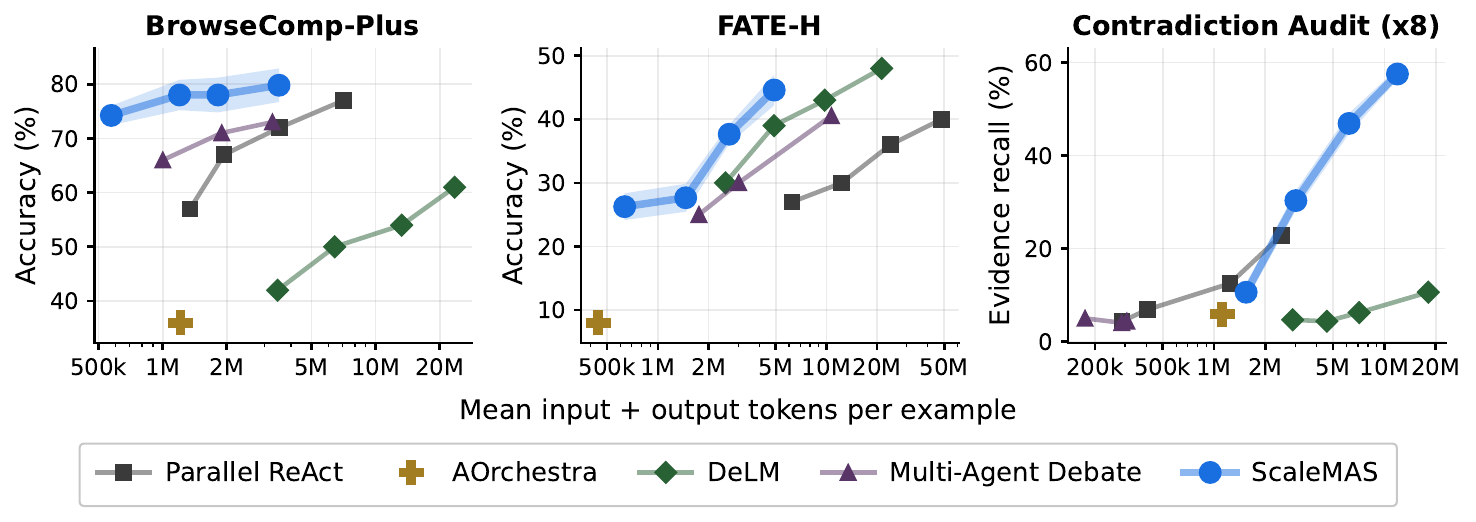}
        \caption{\textbf{Scaling study with Claude-5-Sonnet.} Task performance v.s. inference compute per example (log scale) on BrowseComp-Plus, FATE, and Contradiction Audit. Each point is one group size, scaled from 1 to 2, 4, and 8 agents. At matched compute, \ours outperforms every baseline, and its lead on Contradiction Audit widens as the group grows.}
        \label{fig:performance_sonnet}
    \end{minipage}\hfill
    \begin{minipage}[t]{0.295\linewidth}
        \centering
        \includegraphics[height=\figrowh]{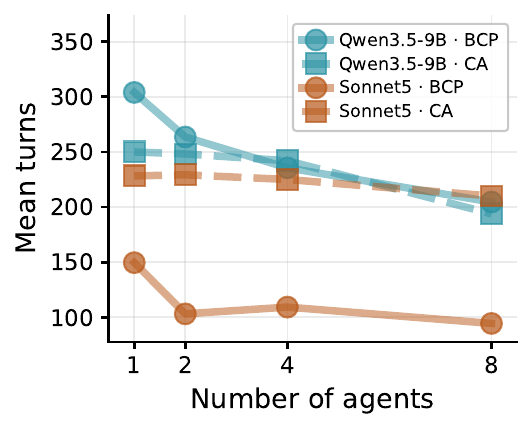}
        \caption{\textbf{Number of turns}, averaged over agents and examples, on BrowseComp-Plus (BCP) and Contradiction Audit \texttt{x8} (CA) with both LLMs.}
        \label{fig:latency}
    \end{minipage}
    \vspace{-10pt}
\end{figure}

\paragraph{Performance grows with group size.}
Figures~\ref{fig:performance} and~\ref{fig:performance_sonnet} plot performance against
the inference compute per example, one point per group size. With Qwen3.5-9B, scaling
\ours from one agent to sixteen raises performance on every benchmark, from $38.1\%$ to
$65.3\%$ on BrowseComp-Plus, from $24.0\%$ to $56.0\%$ on FATE-M, and from $4.1\%$ to
$29.4\%$ on Contradiction Audit \texttt{x4}. On SWE-bench Verified, it levels off after
eight agents, where parallel ReAct saturates as well, which points to the limit of the
LLM's capability. 
Parallel ReAct also improves with group size but stays below \ours at every group size on all four benchmarks. Part of this gap is present with a single agent, where the system degrades to a single agent with a private substrate; \Cref{tab:communication} separates this effect from the benefit of reading one another's messages. 

The baselines behave differently:
on BrowseComp-Plus and Contradiction Audit, adding agents to MAD or DeLM helps little
and sometimes hurts, and neither exceeds $7\%$ on Contradiction Audit at any group
size. Both methods deliver the group's outputs into each agent's context, so adding
agents enlarges the reading burden of every member. Experiments with Claude-5-Sonnet
show similar patterns (\Cref{fig:performance_sonnet}), most visibly on Contradiction
Audit \texttt{x8}, where \ours rises from $10.6\%$ with one agent to $57.5\%$ with eight.

\paragraph{Scaling is more compute-efficient.}
At matched inference compute, \ours is ahead of baselines on every benchmark
(Figures~\ref{fig:performance} and~\ref{fig:performance_sonnet}). For example, with
Qwen3.5-9B (\Cref{fig:performance}), at a similar compute of about $20$M tokens per
example on BrowseComp-Plus, 8 \ours agents reach $60.0\%$ while 16 DeLM agents reach
$26.0\%$; and on SWE-bench Verified, 4 \ours agents reach the same $80.0\%$ as 8 ReAct
agents with about half the tokens ($15.3$M against $28.3$M). The difference grows with
the group: independent ReAct agents repeat each other's searching and reading, so their
tokens per example roughly double with every doubling of the group, whereas on
BrowseComp-Plus and Contradiction Audit the tokens used by \ours grow more slowly than
its size, because agents reuse what others have
already published instead of rediscovering it. The pattern holds with Claude-5-Sonnet:
at about $3.5$M tokens per example on BrowseComp-Plus, eight \ours agents reach
$79.8\%$, against $73.0\%$ for eight MAD agents and $72.0\%$ for four ReAct agents.
Adding agents to MAD, in contrast, increases tokens without improving accuracy: on
BrowseComp-Plus with Qwen3.5-9B, its tokens per example grow from $5.6$M with two
agents to $48.5$M with sixteen, while its accuracy stays between $28\%$ and $32\%$.

\paragraph{Agents take fewer turns in larger groups.}
\Cref{fig:latency} reports the mean number of turns an agent takes before it submits.
It falls as the group grows, from $304$ with one agent to $205$ with eight on
BrowseComp-Plus with Qwen3.5-9B and from $150$ to $94$ with Claude-5-Sonnet, and it also
falls on Contradiction Audit with both backbones. The total work of the group therefore
grows more slowly than its size: eight Qwen3.5-9B agents on BrowseComp-Plus take about
$5.4$ times as many turns in total as a single agent. One reason is that an \ours agent
that finds the result it needs already published can build on it instead of deriving
it again; Section~\ref{sec:analysis} further tests whether correct answers indeed draw
on other agents' work. The turns an agent saves are
mostly calls to the task's own tools: with Claude-5-Sonnet on BrowseComp-Plus, an agent
calls them $80$ times per example in a group of eight against $134$ times alone. On
Contradiction Audit \texttt{x8}, these calls fall from $131$ to $98$ while the agent's
searches of the datastore rise from $12$ to $65$, so agents in larger groups explore the
task less on their own and read more of what the others have found.

\subsection{Analysis}
\label{sec:analysis}
\begin{figure}[t]
    \centering
    \newlength{\figrowha}\setlength{\figrowha}{0.2452\textwidth}
    \begin{minipage}[t]{0.462\linewidth}
        \centering
        \includegraphics[height=\figrowha]{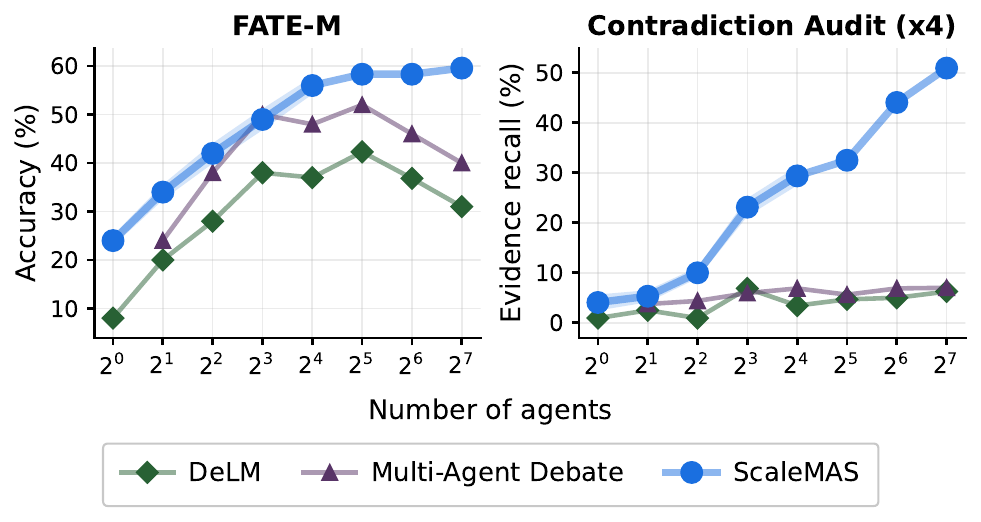}
        \caption{\textbf{Scaling to 128 agents} with Qwen3.5-9B on FATE-M and Contradiction Audit \texttt{x4}, comparing \ours to other multi-agent systems: DeLM and MAD.}
        \label{fig:scaling}
    \end{minipage}\hfill
    \begin{minipage}[t]{0.509\linewidth}
        \centering
        \includegraphics[height=\figrowha]{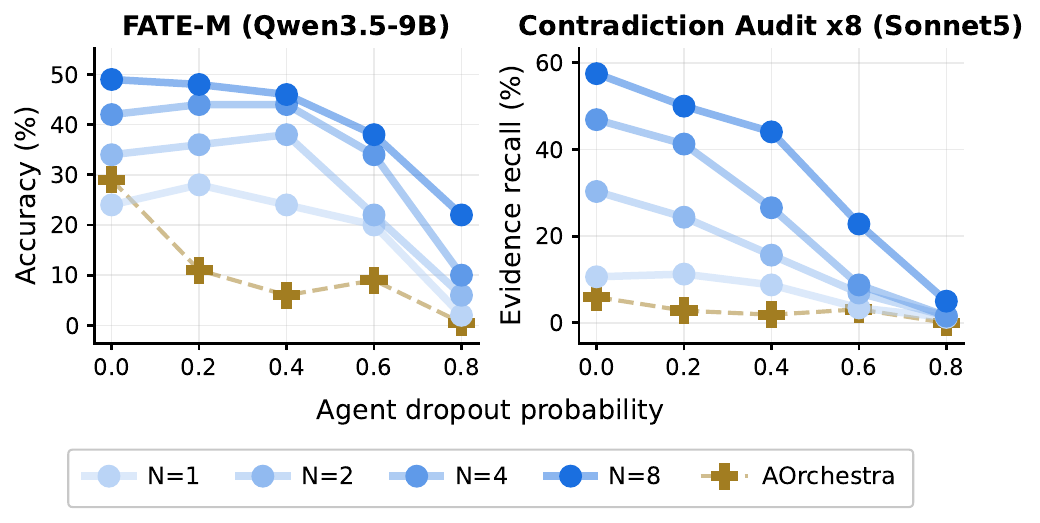}
        \caption{\textbf{Resilience to unreliable membership.} Each agent drops out of each step independently with probability $p$. FATE-M with Qwen3.5-9B and Contradiction Audit \texttt{x8} with Claude-5-Sonnet, 1 to 8 agents, AOrchestra for reference.}
        \label{fig:async}
    \end{minipage}
    \vspace{-10pt}
\end{figure}

\paragraph{Scaling continues to 128 agents.}
\Cref{fig:scaling} extends the sweep to $N=128$ agents with Qwen3.5-9B. On FATE-M,
\ours makes most of its improvement by $N=16$ and then levels off just below $60\%$,
whereas MAD peaks at $N=32$ and declines afterwards, and DeLM ends lower at $N=128$
than at $N=8$. On Contradiction Audit \texttt{x4}, \ours improves at every doubling and
reaches $50.9\%$ at $N=128$, while DeLM and MAD never exceed $7\%$. The baselines thus
show the diminishing or negative returns that prior studies report when multi-agent
systems grow~\citep{kim2026capable},
which we attribute to their unbounded reads. In turn, \ours keeps agents' reads lightweight and turns additional agents into higher performance.

\begin{figure}[t]
    \centering
    \newlength{\figrowhc}\setlength{\figrowhc}{0.272\textwidth}
    \begin{minipage}[t]{0.625\linewidth}
        \centering
        \includegraphics[height=\figrowhc]{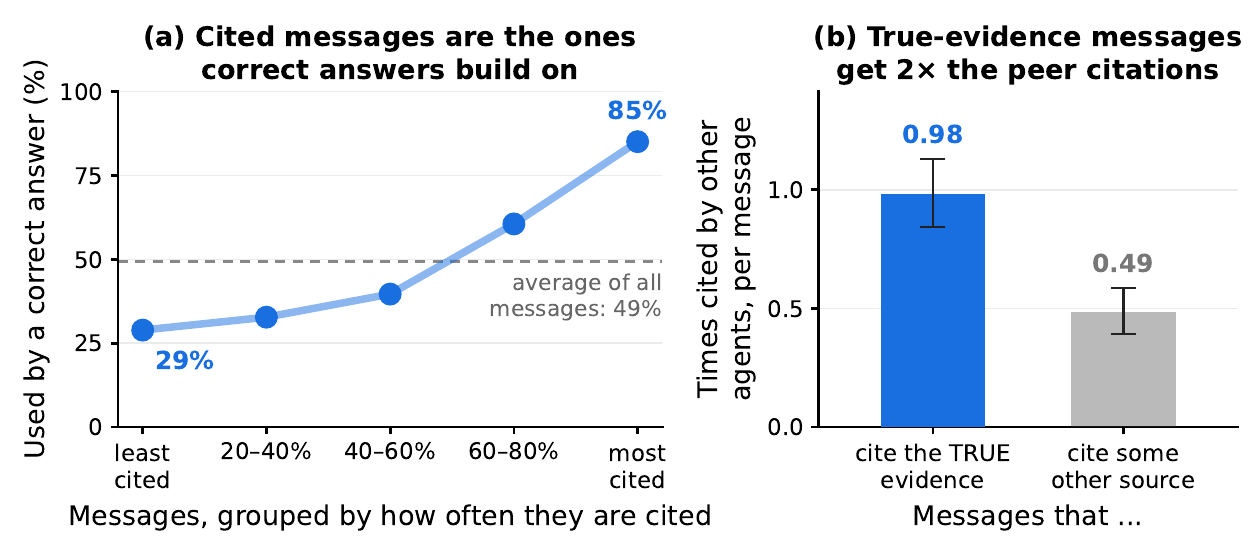}
        \caption{\textbf{Peer citations curate the datastore} (BrowseComp-Plus, all \ours runs). \textbf{(a)} Share of messages on the evidence chain of a correct answer, by fifth of within-run citation count, over the $413$ examples whose correct answer cites a message; dashed: all messages. \textbf{(b)} Mean citations from other agents for source-citing messages that name a gold evidence document or another document, weighting examples equally. Error bars: $95\%$ bootstrap CIs.}
        \label{fig:citation}
    \end{minipage}\hfill
    \begin{minipage}[t]{0.355\linewidth}
        \centering
        \includegraphics[height=\figrowhc]{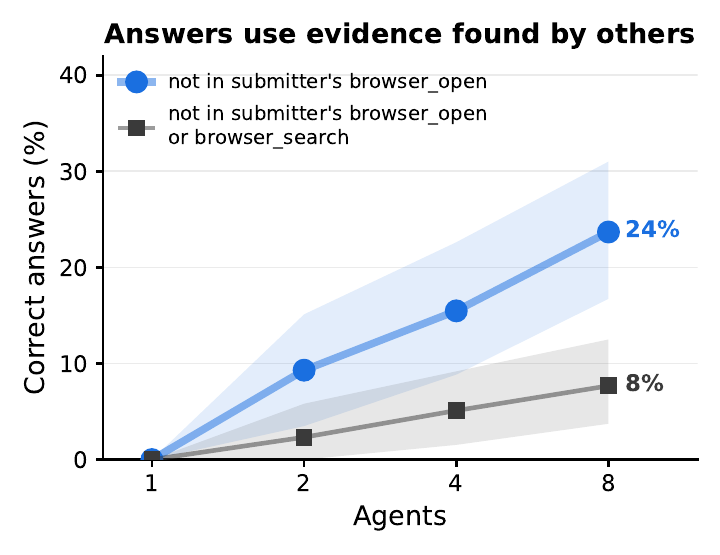}
        \caption{\textbf{Correct answers use evidence found by peers} (BrowseComp-Plus, Qwen3.5-9B). Share of correct answers whose citation chain holds other's message naming a gold document the submitter never opened, or never saw in its searches either.}
        \label{fig:assembly}
    \end{minipage}
    \vspace{-15pt}
\end{figure}

\paragraph{Peer citations automatically curate the datastore.}
The substrate never judges the quality of a message; the only quality-related signal
is which messages the agents choose to cite. We test whether this signal is
informative on all BrowseComp-Plus runs of \ours, using two complementary
criteria. The first is \emph{use}: whether a message lies on the evidence chain of a
correct final answer, that is, whether the answer cites it directly or through a chain
of references. Ranking the messages within each run by citation count, the fraction on
such a chain increases monotonically from $29\%$ in the least-cited quintile to $85\%$
in the most-cited (\Cref{fig:citation}a).
Because the evidence chain itself follows citations, this measure reflects connectivity as well as relevance.

The second criterion is independent of any
answer: whether a message names a gold evidence document annotated by BrowseComp-Plus.
Among messages that name a source, those naming gold evidence receive about twice as
many citations from other agents as those naming other documents
(\Cref{fig:citation}b). These results show that, without any central ranking, peer
citations can effectively single out the messages that matter: those that final answers build on, and
those that hold genuine evidence.

\paragraph{Correct answers are assembled across agents.}
Using BrowseComp-Plus runs with Qwen3.5-9B, we ask whether a
correct answer's evidence chain contains a message, written by another agent, that
names a gold document the submitting agent never opened itself. The share of correct
answers built this way grows with the group, from $9\%$ with two agents to $24\%$ with
eight, and the trend holds under a stricter criterion that also excludes documents
the submitter saw in its own search results (\Cref{fig:assembly}). This indicates that a division of labor emerges although no
agent is ever assigned a sub-task: agents increasingly answer by combining evidence
that others found, and more correct answers rest on it as the group grows.

\paragraph{Individual failures do not break the group.}
\Cref{fig:robustness} adds one agent that fails on every step (i.e. it always returns error messages), holding the number of
healthy agents fixed. For \ours, a group with one failing agent scores about as well
as its healthy agents alone; with seven healthy agents, for instance, both reach
$19.4\%$. This is because \ours does not assign jobs to agents, and does not require an agent to finish
anything: a failing agent simply publishes nothing, while the healthy agents keep
finding and building on each other's messages, so the group loses only the failing
agent's accumulated context. DeLM, in contrast, falls below $1\%$ when one of its members fails. This may be partially due to its task decomposition design, which could be susceptible to individual failures when the errors are injected in the decomposition step.

\paragraph{Agents can leave and rejoin at any time.}
In \Cref{fig:async}, each agent drops out of each step independently with probability
$p$, so agents keep leaving and returning over the course of a run; at $p=0.8$, an agent
is absent from four of every five steps on average and the group's membership changes
constantly. \ours keeps working throughout: even at $p=0.8$, eight agents still solve
$22.0\%$ of FATE-M problems, and on Contradiction Audit \texttt{x8}, eight agents that
drop out of two of every five steps still score higher than every baseline without
dropout. This is because nothing in \ours waits on a particular agent, and what an agent has published
stays in the datastore while it is away, so a returning agent picks up the group's
progress and continues. Larger groups absorb dropout better (at $p=0.8$ on FATE-M,
$22.0\%$ with eight agents against $2.0\%$ with one), whereas AOrchestra, which routes
every result through a single coordinator, loses more than half of its score already
at $p=0.2$ because it cannot tolerate the loss of the coordinator.

\subsection{Ablation Studies}
\label{sec:ablations}

\begin{figure}[t]
    \centering
    \newlength{\figrowhb}\setlength{\figrowhb}{0.212\textwidth}
    \begin{minipage}[t]{0.604\linewidth}
        \centering
        \includegraphics[height=\figrowhb]{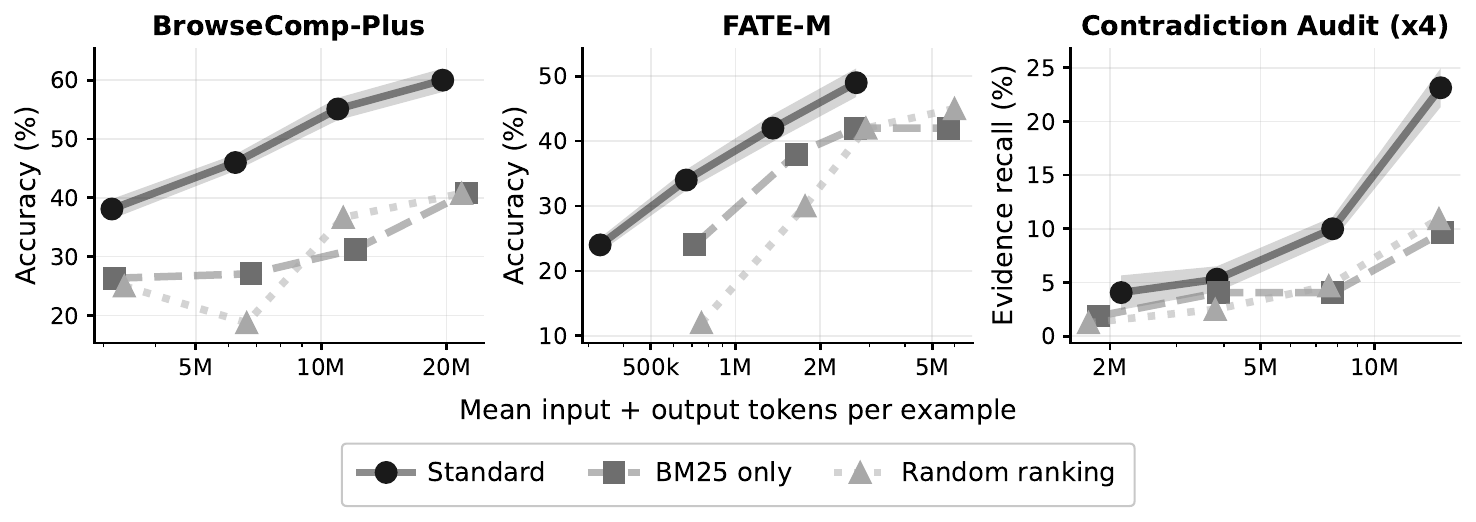}
        \caption{\textbf{Ablating citation weight in search ranking.} Task performance versus mean tokens per example with Qwen3.5-9B, ranking by lexical relevance plus the citation weight (standard), by BM25 alone, or at random. Each point is a group size from 1 to 8.}
        \label{fig:strength}
    \end{minipage}\hfill
    \begin{minipage}[t]{0.366\linewidth}
        \centering
        \includegraphics[height=\figrowhb]{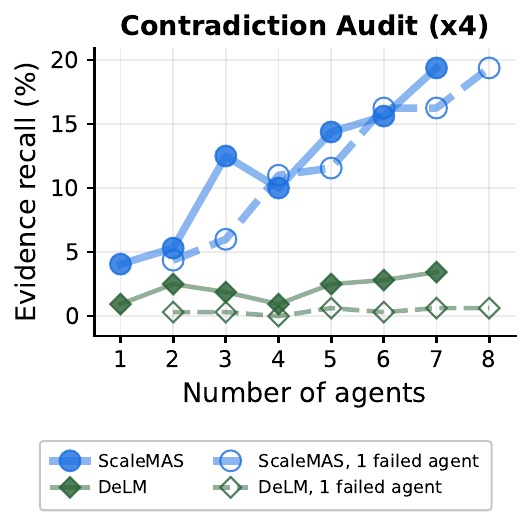}
        \caption{\textbf{One permanently failing agent} on Contradiction Audit \texttt{x4} with Qwen3.5-9B. Dashed: one agent always fails; solid: all agents are healthy.}
        \label{fig:robustness}
    \end{minipage}
\end{figure}

\paragraph{Does the citation weight improve what search returns?}
\Cref{fig:strength} compares the default ranking, which adds the citation weight to
lexical relevance, with BM25 alone and with a random order over matching messages.
Removing the citation weight lowers accuracy at every group size on BrowseComp-Plus
and Contradiction Audit \texttt{x4} without reducing tokens read: at $N=8$, $60.0\%$
against $40.8\%$ and $23.1\%$ against $9.7\%$. On FATE-M the two stay within four
points up to $N=4$ and separate at $N=8$ ($49.0\%$ against $42.0\%$), where BM25
alone also reads twice the tokens ($5.7$M against $2.7$M). A random ranking does as well
as BM25 alone at $N=8$ on all three benchmarks, so
lexical relevance alone recovers little of what the citation weight adds.

\paragraph{Is the improvement from communication, or from running more agents?}
\Cref{tab:communication} holds the group at four agents and removes communication in
two steps. A private substrate, partitioned by author so that each agent reads only
its own messages, lowers accuracy on every benchmark, by $8.3$ points on
BrowseComp-Plus and $8.4$ on SWE-bench Verified. Removing the datastore, which leaves
four independent ReAct agents, lowers it further on three of the four benchmarks
(on Contradiction Audit, $8.1\%$ against $6.6\%$). Reading one another's messages
accounts for about half or more of the improvement over independent sampling.

\begin{table}[t]
\centering
\small
\newcommand{\subrow}{\hspace{0.5em}\begin{picture}(7,6)(0,0)\put(0.5,6.5){\line(0,-1){4}}\put(0.5,2.5){\vector(1,0){6}}\end{picture}\hspace{0.3em}}
\caption{\textbf{Ablating cross-agent communication.} Performance (\%) with four Qwen3.5-9B agents. \emph{Private substrate} partitions the datastore by author, so each agent reads only its own messages; \emph{w/o substrate} runs independent ReAct agents.}
\label{tab:communication}
\begin{tabular}{lcccc}
\toprule
 & \textbf{BrowseComp-Plus} & \textbf{FATE-M} & \begin{tabular}[c]{@{}c@{}}\textbf{SWE-bench}\textbf{Verified}\end{tabular} & \begin{tabular}[c]{@{}c@{}}\textbf{Contradiction}\textbf{Audit (\texttt{x4})}\end{tabular} \\
\midrule
\ours & \textbf{55.1} & \textbf{42.0} & \textbf{80.0} & \textbf{10.0} \\
\subrow private substrate & 46.8 & 38.0 & 71.6 & 6.6 \\
\subrow w/o substrate (ReAct) & 38.0 & 36.0 & 68.0 & 8.1 \\
\bottomrule
\end{tabular}
\end{table}

\section{Related Work}

LLM-based multi-agent systems revisit three coordination structures from classic
distributed problem solving: agents that address one another directly, a designated
agent that mediates, and a shared blackboard. Agents address
one another directly in  \citet{du2024debate} and along the edges of a topology in \citet{qian2025macnet}; a designated agent mediates in \citet{fourney2024magentic} and \citet{ruan2026aorchestra}; and the
blackboard has been revived repeatedly, with \citet{han2025exploring} removing direct contact between agents
entirely and \citet{mao2026shared} placing
typed notes in a context every agent reads. A further line asks which agents
should be connected: \citet{zhuge2024language} optimizes
graph connectivity together with node prompts, and  \citet{zhang2024g} generates a
task-specific topology. These methods change who is connected to whom. What
arrives in a reader's context is still chosen by whoever sends along those edges,
so the question of who selects is left where these structures put it. We discuss
earlier work and further related lines in \Cref{app:rw}.

\section{Conclusion}
\label{sec:conclusion}
We identified three problems that grow with the number of agents: the reading burden
on each agent, the dilution of useful shared work of unknown quality,
and fragility to failing agents. \ours addresses them with an agent substrate: agents interact only through a shared datastore of linked
messages, every read returns at most $k$ messages ranked by relevance and by how often
they are cited, and messages persist after their producers leave. Across four
benchmarks and two backbones, \ours improves with group size up to eight agents on
every benchmark, and with Qwen3.5-9B up to sixteen on three of the four, while reaching
higher performance than message-passing, coordinator-based, and blackboard-style
baselines at matched inference compute. Our analysis shows the mechanisms at work: peer
citations concentrate attention on messages that hold genuine evidence, correct answers
increasingly draw on evidence found by other agents, and the group tolerates both a
permanently failing agent and agents that drop out at any step.

\paragraph{Limitations and future work.}
Due to limited compute, our experiments scale groups up to $128$ agents; larger groups
are left to future work. \ours is also used only at inference time, and training agents
to coordinate through the substrate is another direction for future work.


\clearpage
\newpage
\bibliographystyle{assets/plainnat}
\bibliography{ref}

\clearpage
\newpage

\beginappendix
\startcontents[appendix]
\section*{Content}
\definecolor{appdarkblue}{RGB}{23,55,130}
{\hypersetup{linkcolor=appdarkblue}\color{appdarkblue}%
\printcontents[appendix]{}{1}{\setcounter{tocdepth}{2}}}
\clearpage

\section{Extended Related Work}
\label{app:rw}
\paragraph{Cooperative problem solving and coordination.}
How a group of problem solvers should divide work and share partial results is an
old question, surveyed for the pre-LLM era in \citet{durfee1989trends}. The three structural answers this paper builds on were fixed in
that period. The blackboard metaphor appears in  \citet{newell1962problems}, which already describes workers who
each read everything on a shared surface, and was built in \citet{erman1980hearsay}, whose
knowledge sources are activated by what appears on the blackboard rather than by
calls from one another. Direct addressing was put to work for task allocation in
\citet{smith1980contract}, whose nodes exchange messages and share no memory, while \citet{gelernter1985generative} gives a shared medium in which writers do
not name readers. The communication complexity of such structures was studied in parallel, from  \citet{yao1979complexity} to \citet{dolev1989multiparty}, in terms of bits on a channel. The readers that
literature assumes can scan a growing shared surface with little effort, so the amount of
reading was never the binding constraint. It becomes one when the reader is a language model.

\paragraph{Reducing what agents read.}
A second line accepts the delivery structure and makes each delivery smaller.
 \citet{zeng2025s2}
propagates only exchanges it judges novel, and  \citet{chen2025optima} trains agents to communicate
more compactly; a parallel effort operates inside a single agent, compressing or
evicting accumulated history to keep a long run within the context window \citep{wu2026contextbudget, semenov2026beyond}. Both
reduce the size of what has already been routed to an agent. The bound studied
here applies one step earlier, to how much is routed at all, and is set by the
reader rather than by the sender. The two are complementary: every method in our
experiments, \ours included, manages its own history in the same way (\Cref{app:harness}).

\paragraph{Retrieval, memory, and information access.}
Shared stores for language agents are usually built to persist:  \citet{zhong2024memorybank} and \citet{park2023generative} maintain per-agent stores across
sessions, and \citet{xu2026mem} links notes and rewrites
older ones. The datastore studied here is not of this kind: it is created for one task
and discarded with it, and exists so that agents working at the same time can
reach each other's results. Retrieval by an agent has also been studied where the
corpus is fixed in advance \citep{singh2025agentic}; here the corpus is the group's own output, so what an agent can
find at any moment is what the others have published by that moment. The
distinction our substrate rests on is older than either literature:  \citet{belkin1992information} separates
systems in which profiles decide what reaches a reader from systems in which a
reader expresses a need and initiates retrieval. The three families of
\Cref{sec:intro} are of the first kind; the design of
\Cref{sec:method} is of the second, and takes its two access modes from the
tradition of \citet{bush1945we} and  \citet{BernersLee1990InformationMA}, with
ranking informed by reference structure as in  \citet{brin1998anatomy}.

\section{Extended Details on Contradiction Audit}
\label{app:ca}

This section describes details about how Contradiction Audit (Section~\ref{sec:contradiction-audit}) is
built and scored. The pipeline has four stages: collecting documentation, splitting it
into snippets, injecting contradictions, and deriving difficulty levels.

\subsection{Corpora}
\label{app:ca-corpora}

The benchmark covers sixteen open-source projects: CPython, Docker, Godot, JAX,
Kubernetes, the Linux kernel, LLVM, NumPy, pandas, PyTorch, Ray, SciPy, scikit-learn,
SQLAlchemy, SymPy, and Hugging Face Transformers. From the published documentation of
each project we keep only Markdown and reStructuredText files. Files that disagree with
the rest of the documentation by design are excluded, since they would contain
contradictions we did not inject: release notes, changelogs and ``what's new'' pages;
deprecation notices, roadmaps and to-do pages, which describe changing state; fragments
that are transcluded verbatim into other pages; documentation templates; and
contributor, governance and license pages. For a few projects we also exclude
version-specific API dumps and migration guides. \Cref{tab:ca-corpora} lists the size of
each corpus.

\begin{table}[h]
\centering
\small
\caption{The sixteen documentation corpora of Contradiction Audit, with the number of
files kept after filtering and the number of snippets in the full corpus.}
\label{tab:ca-corpora}
\begin{tabular}{lrr@{\hspace{2.5em}}lrr}
\toprule
Project & Files & Snippets & Project & Files & Snippets \\
\midrule
CPython      & 507     & 2{,}806 & pandas       & 78  & 392 \\
Docker       & 930     & 2{,}074 & PyTorch      & 245 & 597 \\
Godot        & 1{,}586 & 4{,}512 & Ray          & 534 & 1{,}347 \\
JAX          & 233     & 920     & SciPy        & 339 & 635 \\
Kubernetes   & 1{,}630 & 3{,}115 & scikit-learn & 102 & 475 \\
Linux kernel & 3{,}433 & 8{,}319 & SQLAlchemy   & 152 & 583 \\
LLVM         & 1{,}229 & 2{,}663 & SymPy        & 317 & 732 \\
NumPy        & 199     & 573     & Transformers & 726 & 1{,}507 \\
\midrule
\multicolumn{6}{l}{Total: $12{,}240$ files and $31{,}250$ snippets.} \\
\bottomrule
\end{tabular}
\end{table}

\subsection{Splitting into snippets}
\label{app:ca-snippets}

Each file is split into paragraphs, that is, runs of non-blank lines, and the paragraphs
are packed greedily into snippets of about $1024$ tokens (\texttt{tiktoken}
\texttt{o200k\_base}). A paragraph is never split, so every snippet ends at a paragraph
boundary and may run slightly over the target to complete one; a file containing a
single paragraph longer than twice the target is skipped. Snippets keep the order of
their files, and each snippet begins with a line naming the file it comes from, which
the agents see.

\subsection{Injecting contradictions}
\label{app:ca-inject}

Contradictions are injected into pairs of snippets by GPT-5.6, one pair at a time, in
five steps.

\paragraph{Candidate pairs.}
Random pairs of snippets almost never discuss the same subject, so candidates are drawn
from lexical overlap. A term is a lowercased word of at least four characters or one
containing a digit; a pair is a candidate if the two snippets share at least two terms
that each occur in at most eight snippets of the corpus. Candidates are visited in
decreasing order of shared terms, with ties broken by a fixed random seed.

\paragraph{Shared subject.}
The model is shown the two snippets inside their full files, so that names and
references can be resolved, and asked for up to three subjects that both snippets
discuss and that are central to what their files document, rather than passing
mentions. Pairs without such a subject are skipped.

\paragraph{Edit.}
The model rewrites the prose of one of the two snippets so that it can no longer be
true together with the other, and explains why the two now conflict. It may only edit
natural-language prose, never code, tables, list entries or headings, and must copy the
rest of the snippet unchanged.

\paragraph{Mechanical checks.}
The edited snippet is compared with the original. The edit is rejected if it changes
more than $30\%$ of the snippet, if its changes are spread over more than eight
separate places, or if it changes only formatting. Each sentence containing an edit
must also occur exactly once in the corpus, so that the injected claim can be located
without ambiguity.

\paragraph{Judges.}
Four independent GPT-5.6 judges then check that the edited snippet reads naturally on
its own, that the contradiction is not trivial, so that a model cannot find it from
memorized knowledge of the project, that the edited snippet does not contradict
itself, and that the two snippets truly cannot both be true. A contradiction is kept
only if all four judges accept it.

\medskip\noindent
The walk over candidates stops once each corpus holds twenty contradictions: ten whose
two snippets come from the same file and ten whose snippets come from different files.
Every snippet takes part in at most one contradiction, so the twenty contradictions
occupy forty distinct snippets.

\subsection{Difficulty levels}
\label{app:ca-levels}

Level \texttt{x}$n$ keeps the forty contradiction snippets and adds $40(n-1)$
distractor snippets drawn at random with a fixed seed, for $40n$ snippets in total.
Distractors are drawn first from snippets that form a whole file on their own, so that
the reduced corpus consists of complete files as far as possible. The kept snippets
stay in their original corpus order and are renumbered from zero. The contradictions
are identical at every level, so a higher level differs only in how much unrelated
text surrounds them. The main experiments use \texttt{x4} ($160$ snippets per corpus)
with Qwen3.5-9B and \texttt{x8} ($320$ snippets) with Claude-5-Sonnet.

\subsection{Tools and reporting}
\label{app:ca-tools}

Every method receives the same two corpus tools. \texttt{fetch\_snippet} returns
the full text of a snippet given its index. \texttt{search\_snippet} is an exact, case-insensitive text search over the corpus, which
returns up to ten matching snippets with the surrounding context of each match. Each agent is told the number of snippets and a starting position; the starting
positions of a group are randomly set. A
contradiction is reported as a pair of snippet indices with a short reason, and an
agent's final answer is the list of pairs it is confident of. In \ours, a reported
contradiction is also published as a message whose rationale quotes the conflicting
text of both snippets; a quote that is not verbatim text of the cited snippets is
rejected.

\subsection{Scoring}
\label{app:ca-scoring}

A contradiction counts as found when some reported pair includes its edited snippet.
Reported pairs that match no injected contradiction are not penalized. The score of a
run is the fraction of the twenty contradictions found by the pairs its agents report,
and the score of a method is this fraction averaged over the sixteen corpora.

\subsection{Examples}
\label{app:ca-examples}

\Cref{tab:ca-examples-same,tab:ca-examples-cross} show one contradiction from each of
the sixteen corpora: eight whose two snippets come from the same file and eight whose
snippets come from different files. Each excerpt keeps the conflicting sentence and a
little of the text around it, and omits the rest of the snippet. In every pair the
unedited snippet still states the original claim, while the edited snippet now states
one that cannot be true together with it; the last column states the conflict.

\begin{table}[p]
\centering
\footnotesize
\newcommand{\injected}[1]{\textcolor[rgb]{0.55,0.15,0.15}{#1}}
\caption{\textbf{Contradiction Audit examples within one file}, one per corpus. Both snippets come from the file named after the project. The injected text is in red, and \ldots\ marks omitted text.}
\label{tab:ca-examples-same}
\begin{tabular}{@{}p{0.355\linewidth}@{\hspace{0.03\linewidth}}p{0.355\linewidth}@{\hspace{0.03\linewidth}}p{0.23\linewidth}@{}}
\toprule
Unedited snippet & Edited snippet & Why they conflict \\
\midrule
\multicolumn{3}{@{}l}{\textbf{CPython}\quad \texttt{library/\allowbreak{}math.rst}} \\
\raggedright \ldots\ \texttt{atan2(y, x)}: \texttt{atan(y / x)}, in radians \ldots
&
\raggedright \ldots\ \texttt{atan2(y, x)}: \injected{Return \texttt{atan(y / x)}, in degrees. The result is between \texttt{-180} and \texttt{180}.} \ldots
&
\raggedright The function summary gives the result of \texttt{atan2} in radians; its own entry now gives it in degrees. \tabularnewline
\midrule
\multicolumn{3}{@{}l}{\textbf{Kubernetes}\quad \texttt{reference/\allowbreak{}access-authn-authz/\allowbreak{}authentication.md}} \\
\raggedright \texttt{-{}-oidc-issuer-url}: URL of the provider that allows the API server to discover public signing keys. Only URLs that use the \texttt{https://} scheme are accepted.
&
\raggedright \ldots\ \injected{The URL scheme may be either \texttt{http} or \texttt{https}; plain \texttt{http} issuers are accepted.}
&
\raggedright An \texttt{http} issuer URL is rejected in one and accepted in the other. \tabularnewline
\midrule
\multicolumn{3}{@{}l}{\textbf{LLVM}\quad \texttt{ProjectGovernance.md}} \\
\raggedright \ldots\ An individual cannot serve on two \emph{area teams}. Members of an \emph{area team} are elected for 1 year terms.
&
\raggedright \ldots\ \injected{Elections for area teams occur in January every third year, and members elected in these elections serve three-year terms.}
&
\raggedright Area team members serve one-year terms in one and three-year terms in the other. \tabularnewline
\midrule
\multicolumn{3}{@{}l}{\textbf{NumPy}\quad \texttt{reference/\allowbreak{}simd/\allowbreak{}build-options.rst}} \\
\raggedright \ldots\ Specifying features to \texttt{cpu-dispatch} or \texttt{cpu-baseline} doesn't explicitly enable them. Features are detected at compile time, and the maximum available features based on your specified options will be enabled according to toolchain and platform support.
&
\raggedright \texttt{MAX}: \injected{Enables every known feature unconditionally, even when the compiler or platform does not support it.}
&
\raggedright A requested feature is enabled only if the toolchain and platform support it, so the \texttt{MAX} option cannot enable features unconditionally. \tabularnewline
\midrule
\multicolumn{3}{@{}l}{\textbf{pandas}\quad \texttt{development/\allowbreak{}developer.rst}} \\
\raggedright \ldots\ For types other than these, the \texttt{'metadata'} key can be omitted. Implementations can assume \texttt{None} if the key is not present.
&
\raggedright \ldots\ \injected{The \texttt{metadata} field is always required and must never be \texttt{None} nor omitted; it holds an empty dictionary except for:} \ldots
&
\raggedright The \texttt{metadata} key may be omitted for most types in one and must always be present in the other. \tabularnewline
\midrule
\multicolumn{3}{@{}l}{\textbf{SciPy}\quad \texttt{dev/\allowbreak{}api-dev/\allowbreak{}array\_api.rst}} \\
\raggedright \ldots\ The restriction that functions be functionally pure to support the JAX JIT necessitates that input-validation that raises with bad input must be skipped when \texttt{xp} is JAX.
&
\raggedright \ldots\ Functions using the JAX JIT must be functionally pure. They cannot have side effects, cannot mutate data, and their outputs must be determined completely by their inputs. \injected{Raising a Python exception, however, is not considered a side-effect and is fully permitted within a JITed function.}
&
\raggedright Input validation that raises must be skipped under the JIT in one, because raising is not allowed there; the other says raising is fully permitted. \tabularnewline
\midrule
\multicolumn{3}{@{}l}{\textbf{SQLAlchemy}\quad \texttt{dialects/\allowbreak{}postgresql.rst}} \\
\raggedright \ldots\ \texttt{native\_inet\_types=False}. With the above parameter, the \texttt{psycopg}, \texttt{asyncpg} and \texttt{pg8000} dialects will disable the DBAPI's adaptation of these types and will return only strings \ldots
&
\raggedright \ldots\ \injected{To \textbf{normalize all the above DBAPIs to only return strings}, use the \texttt{native\_inet\_types} parameter, passing a value of \texttt{True}:} \ldots
&
\raggedright Only strings are returned with \texttt{native\_inet\_types=False} in one and with \texttt{True} in the other. \tabularnewline
\midrule
\multicolumn{3}{@{}l}{\textbf{SymPy}\quad \texttt{tutorials/\allowbreak{}intro-tutorial/\allowbreak{}manipulation.rst}} \\
\raggedright \ldots\ There is no subtraction class in SymPy. \texttt{x - y} is represented as \texttt{x + -y} \ldots
&
\raggedright \ldots\ \injected{We might expect to see something like \texttt{Div(1, y)}, but unlike subtraction, which has its own class, none exists for division.}
&
\raggedright SymPy has no subtraction class in one, and subtraction has its own class in the other. \tabularnewline
\bottomrule
\end{tabular}
\end{table}

\begin{table}[p]
\centering
\footnotesize
\newcommand{\injected}[1]{\textcolor[rgb]{0.55,0.15,0.15}{#1}}
\caption{\textbf{Contradiction Audit examples across two files}, one per corpus. Each excerpt is headed by the file its snippet comes from. The injected text is in red, and \ldots\ marks omitted text.}
\label{tab:ca-examples-cross}
\begin{tabular}{@{}p{0.355\linewidth}@{\hspace{0.03\linewidth}}p{0.355\linewidth}@{\hspace{0.03\linewidth}}p{0.23\linewidth}@{}}
\toprule
Unedited snippet & Edited snippet & Why they conflict \\
\midrule
\multicolumn{3}{@{}l}{\textbf{Docker}} \\
\raggedright \texttt{guides/\allowbreak{}testcontainers-java-lifecycle.md}\newline \ldots\ Even without explicitly stopping the container in \texttt{@AfterAll}, Testcontainers uses the Ryuk container to clean up containers automatically when the JVM exits.
&
\raggedright \texttt{guides/\allowbreak{}testcontainers-java-getting-started.md}\newline \ldots\ \injected{Note that Testcontainers never removes containers on its own, so you must always stop the container explicitly in \texttt{@AfterAll}; otherwise it keeps running even after the JVM exits.}
&
\raggedright Testcontainers removes containers automatically when the JVM exits in one, and never removes them in the other. \tabularnewline
\midrule
\multicolumn{3}{@{}l}{\textbf{Godot}} \\
\raggedright \texttt{community/\allowbreak{}asset\_library/\allowbreak{}submitting\_to\_assetlib.rst}\newline \ldots\ No \textbf{submodules}, or any submodules must be non-essential. GitHub does not include submodules in the downloaded ZIP file, so if the asset needs the contents of the submodule, your asset won't work.
&
\raggedright \texttt{community/\allowbreak{}asset\_store/\allowbreak{}submitting\_to\_asset\_store.rst}\newline \ldots\ \injected{\textbf{Submodules} are allowed, even essential ones. GitHub includes submodules in the downloaded ZIP file, so if the asset needs the contents of the submodule, it will still work.}
&
\raggedright GitHub's ZIP download leaves out submodules in one and includes them in the other. \tabularnewline
\midrule
\multicolumn{3}{@{}l}{\textbf{JAX}} \\
\raggedright \texttt{export/\allowbreak{}export.md}\newline \ldots\ Note that the rehydrated function does not support any other transformations, e.g., forward-mode AD (jvp), or \texttt{jax.vmap}.
&
\raggedright \texttt{new\_docs/\allowbreak{}501/\allowbreak{}export.md}\newline \ldots\ \injected{Note that the rehydrated function also supports all other transformations, e.g., forward-mode AD (jvp), and \texttt{jax.vmap}.}
&
\raggedright The rehydrated function supports no other transformations in one, and all of them in the other. \tabularnewline
\midrule
\multicolumn{3}{@{}l}{\textbf{Linux kernel}} \\
\raggedright \texttt{filesystems/\allowbreak{}proc.rst}\newline \ldots\ Absolute values of the counters in this line are not accurate because of the failure to allocate memory to track some of the allocations made at this location.
&
\raggedright \texttt{mm/\allowbreak{}allocation-profiling.rst}\newline \ldots\ Runtime info: \texttt{/proc/allocinfo} \injected{Memory needed to track allocations is always reserved up front and can never fail to be allocated, so the absolute values of the counters reported in this file are always exact.}
&
\raggedright The counters in \texttt{/proc/allocinfo} are inaccurate in one and always exact in the other. \tabularnewline
\midrule
\multicolumn{3}{@{}l}{\textbf{PyTorch}} \\
\raggedright \texttt{user\_guide/\allowbreak{}torch\_compiler/\allowbreak{}compile/\allowbreak{}programming\_model.compiler\_disable.md}\newline \ldots\ When the compiler crashes, we will just skip tracing the function and try again later. \textbf{This is not best practice} - it is better to eventually manually add \texttt{disable} annotations as necessary.
&
\raggedright \texttt{user\_guide/\allowbreak{}torch\_compiler/\allowbreak{}torch.compiler\_troubleshooting.md}\newline \ldots\ When the compiler crashes, we will just skip tracing the function and try again later. \injected{This is the recommended best practice - it is better than eventually manually adding disable annotations.}
&
\raggedright Suppressing compiler errors is not best practice in one, and the recommended practice in the other. \tabularnewline
\midrule
\multicolumn{3}{@{}l}{\textbf{Ray}} \\
\raggedright \texttt{serve/\allowbreak{}tutorials/\allowbreak{}stable-diffusion.md}\newline \ldots\ The autoscaling config sets \texttt{min\_replicas} to 0, which means the deployment starts with no \texttt{ObjectDetection} replicas. These replicas spawn only when a request arrives. \ldots
&
\raggedright \texttt{serve/\allowbreak{}tutorials/\allowbreak{}text-classification.md}\newline \ldots\ \injected{The autoscaling config sets \texttt{min\_replicas} to 1, because Ray Serve doesn't support scaling a deployment down to 0 replicas.} \ldots
&
\raggedright Ray Serve can scale a deployment down to zero replicas in one, and can never go below one replica in the other. \tabularnewline
\midrule
\multicolumn{3}{@{}l}{\textbf{scikit-learn}} \\
\raggedright \texttt{datasets/\allowbreak{}sample\_generators.rst}\newline \texttt{make\_blobs} creates a multiclass dataset by allocating each class to one normally-distributed cluster of points. It provides control over the centers and standard deviations of each cluster. This dataset is used to demonstrate clustering.
&
\raggedright \texttt{developers/\allowbreak{}minimal\_reproducer.rst}\newline Similarly to \texttt{make\_classification}, \texttt{sklearn.datasets.make\_blobs} creates multiclass datasets using normally-distributed clusters of points. \injected{It provides no control regarding the centers and standard deviations of each cluster, which are always drawn at random, so it is not suited to demonstrate clustering.}
&
\raggedright \texttt{make\_blobs} lets the user set cluster centers and standard deviations in one, and does not in the other. \tabularnewline
\midrule
\multicolumn{3}{@{}l}{\textbf{Transformers}} \\
\raggedright \texttt{model\_doc/\allowbreak{}matcha.md}\newline \ldots\ MatCha is a model that is trained using \texttt{Pix2Struct} architecture.
&
\raggedright \texttt{model\_doc/\allowbreak{}deplot.md}\newline \ldots\ \injected{DePlot is the only model in the Transformers library that is trained using \texttt{Pix2Struct} architecture.}
&
\raggedright MatCha, also in the library, is trained with the same architecture, so DePlot cannot be the only one. \tabularnewline
\bottomrule
\end{tabular}
\end{table}

\section{Implementation Details}
\subsection{Hyperparameters}
\label{app:hyper}

\Cref{tab:hyper} lists the settings used in all experiments unless stated otherwise.
The citation weight follows Section~\ref{sec:ranking}: a new message starts at weight
$s_0$, each message that cites it moves its weight a fraction $\alpha$ of the way toward
one, and all weights decay by a factor $\gamma$ at every step. Each agent may take a
limited number of actions, where every tool call counts as one action, including a
call that is rejected; a run also ends after a maximum number of steps, at which point
every agent still running submits the best answer it can from the work already done.

\begin{table}[h]
\centering
\small
\caption{Settings of \ours in all experiments unless stated otherwise. The same temperature,
generation limit, and random seeds are shared by every other baseline.}
\label{tab:hyper}
\begin{tabular}{lcccc}
\toprule
Setting & BrowseComp-Plus & \begin{tabular}[c]{@{}c@{}}Contradiction\\Audit\end{tabular} & FATE & \begin{tabular}[c]{@{}c@{}}SWE-bench\\Verified\end{tabular} \\
\midrule
Messages returned per \texttt{search} ($k$) & \multicolumn{4}{c}{$8$} \\
Messages per \texttt{visit} call & \multicolumn{4}{c}{$8$} \\
Retrieval function, BM25 $k_1$ and $b$ & \multicolumn{4}{c}{BM25, $1.5$ and $0.75$} \\
Citation weight $\alpha$, $\gamma$, $s_0$ & \multicolumn{4}{c}{$0.3$, $0.98$, $0$} \\
References per message (maximum) & \multicolumn{4}{c}{$32$} \\
Tokens per message (sub-task and artifact) & $256$ & $256$ & $1{,}024$ & $1{,}024$ \\
\midrule
Actions per agent (maximum) & $300$ & $250$ & $100$ & $300$ \\
Steps per run (maximum) & $400$ & $250$ & $120$ & $400$ \\
Context summarization & yes & yes & yes & no \\
\midrule
Temperature & \multicolumn{4}{c}{$1.0$} \\
Generation limit per call (tokens) & \multicolumn{4}{c}{$16{,}384$} \\
Tool-call format & \multicolumn{4}{c}{native tool calling} \\
Random seed & \multicolumn{4}{c}{$0,1,2$} \\
\bottomrule
\end{tabular}
\end{table}

\subsection{Agent harness}
\label{app:harness}
\ours shares the identical agent harness with all baselines: ReAct, DeLM, AOrchestra, and MAD. For Contradiction Audit, FATE, and BrowseComp-Plus, the agent summarizes its own history periodically: once the history reaches
ten turns, all but the three most recent turns are replaced by a summary of at most
$1{,}024$ tokens written by the same model, and an earlier summary is rewritten
together with the newly summarized turns. The mini-SWE-agent harness used on SWE-bench
Verified keeps the full history instead.

\subsection{Evaluation details}
\label{app:eval}
For the three third-party benchmarks, we report Pass@N accuracy (where N is the number of agents), following other multi-agent systems work \citep{park2026scalingdiscoverytesttimecommunication, mao2026shared}. This evaluation protocol is shared across all methods we tested. Pass@N measures what the group finds rather than which answer it would commit to; we treat answer selection as orthogonal, and single out the benefit of coordination by comparing with parallel ReAct agents.

Every method submits a final answer when its action limit is reached: ReAct and MAD through the forced final step of their agent loop, DeLM and AOrchestra by submitting the answer recorded so far. Each baseline follows its released design, adapted to the shared benchmark adapter.

\paragraph{BrowseComp-Plus.}
We use the questions and the document corpus of BrowseComp-Plus~\citep{chen2025browsecomp},
served through the search service and browser tools of OpenResearcher. Agents have
three tools: \texttt{browser\_search} returns the top $20$ documents of a BM25 search
over the corpus; \texttt{browser\_open} opens a document or a link on the current page
and shows a window of about $1{,}024$ tokens, which the agent can move; and
\texttt{browser\_find} searches for a text pattern in an open page. Each agent's final
answer is graded by an LLM judge (Qwen3.5-flash) with OpenResearcher's grading prompt,
which extracts the final answer from the response and compares it with the gold
answer. Experiments use the first $100$ questions.

\paragraph{FATE.}
FATE~\citep{jiang2026fate} poses problems in abstract algebra as formal statements in
Lean~4 with Mathlib, at increasing levels of difficulty; we use FATE-M ($150$ problems)
and FATE-H ($100$ problems). Agents have two tools: \texttt{search\_mathlib} returns the
$10$ best-matching declarations from an index of Mathlib, and \texttt{check\_lean}
compiles a candidate proof and returns the compiler's messages, at most $10$ times per
agent with a limit of $900$ seconds per compilation. A final proof is compiled with
Lean~4 (v4.28.0) and counts as correct only if it proves the stated theorem, compiles
without errors, and contains no \texttt{sorry}. Experiments use the first $100$ questions of FATE-M, and use all questions of FATE-H.

\paragraph{SWE-bench Verified.}
We run SWE-bench Verified~\citep{jimenez2024swe} on the mini-SWE-agent
scaffold~\citep{yang2024swe}. Each agent works in its own container holding the
repository at the base commit of the issue, runs shell commands with a limit of $60$
seconds per command, and submits the diff of its changes as a patch. A patch is graded
with the benchmark's tests: it is applied to a fresh copy of the repository, and the
instance counts as resolved only if all tests that the fix must make pass
(\texttt{FAIL\_TO\_PASS}) and all tests that must keep passing (\texttt{PASS\_TO\_PASS})
pass. Experiments use the first $100$ instances of the test split.

\subsection{Prompts}
\label{app:prompts}

The part of the system prompt that describes the datastore and its operations is shared by every benchmark. In it, \texttt{publish}, \texttt{search}, and \texttt{visit} appear as the tools \texttt{publish\_message}, \texttt{search\_datastore}, and \texttt{visit\_messages}:

\begin{lstlisting}[basicstyle=\ttfamily\scriptsize, breaklines=true, breakatwhitespace=true, breakautoindent=true, columns=fullflexible, keepspaces=true, frame=single, framesep=5pt, xleftmargin=5pt, xrightmargin=5pt, upquote=true, literate={"}{{\char34}}1]
You are working on a multi-step task across many sessions.

## The datastore: your record of the work

The datastore is a pile of messages holding everything already done on the input task.
Every message in it is an earlier work. You resume from them rather than start over.

- `publish_message(sub_task, artifact, rationale, refs)` is the ONLY way anything
  you do leaves a trace. Work you finish but never write down is lost.
- Every message records work ALREADY DONE, in three parts:
  - **sub_task** -- the piece of work you did, in the PRESENT PERFECT
    (e.g. "I've processed Resource8", "I've ruled out the second candidate"),
    stated in the words you would later search for to find it again;
  - **artifact** -- what the work produced: the result that can be picked up and
    built on without redoing it;
  - **rationale** -- BRIEFLY, how you got there, NAMING each of your refs and
    what it contributed (e.g. "Message12 gave the bound").
  Plus refs (pointers to other messages or to the resources your tools expose),
  and an id. Messages are never edited or deleted -- you only add new ones.
- Messages carry sequential names -- e.g. Message0, Message23 -- usable in refs.
- Any unit of work can be a message -- a digest of something you read, a verification of
  an earlier message, a consolidation of several, a piece of the deliverable. So
  can *raising* something: the `sub_task` is "Asked an open question", and the
  artifact is the question itself.
- MOST of what you write should be INTERMEDIATE work: a partial
  result, a lead worth taking, a dead end, something you ruled out, a step that
  turned out to be wrong, etc.
  Write these as messages AS YOU GO rather than saving up a finished answer.

## How to work

1. READ BEFORE YOU WRITE. BEFORE starting a piece of work, search for it -- the
   wording of the sub-task you are about to do, and the name of anything you
   would work on. If a message already declares that work, it is DONE: build on
   it and move to uncovered ground instead of repeating it.
   TWO READ PATHS:
   1) `search_datastore(<query>)` finds messages by their WORDING -- use it when you do not yet know which messages matter.
   2) `visit_messages(<message ids>)` fetches messages whose names
   you ALREADY know: from a message's refs, from a search hit that cited
   something, from anywhere an id was put in front of you.
2. GROUND EVERYTHING, AND SAY HOW. Refs declare what a message builds on -- e.g.
   "refs": ["Message23"] is a message building on message 23. A Message ref merges your
   work into an existing trail instead of starting a new one; a resource ref
   records what the message draws on so the source can be checked. Only reference names you have actually seen.
   IF YOU USED A MESSAGE, REFERENCE IT. If an earlier message's result saved you a step, sent you down
   a route, or you checked, corrected, extended, or contradicted it, put it in
   your refs. Two rules make a citation checkable, and BREAKING EITHER ONE REJECTS THE
   WHOLE MESSAGE -- it is not written, and you are told exactly which ref was at
   fault:
   - **IT MUST EXIST.** A ref names something you have ACTUALLY SEEN.
   - **YOUR RATIONALE MUST NAME IT, with what it contributed** -- e.g. "Message23 ruled
     out the direct route, so this takes the inductive one", "Input7 supplied
     the figure this checks".
3. WRITE TO BE FOUND. Your message is retrieved by its THREE FIELDS, nothing
   else, and messages containing a query's **exact wording**
   score far above messages merely sharing some of its words: use the exact,
   specific words (and names) that would be searched for later. This matters most
   in `sub_task`: it is what stops the same work being done twice. Keep
   `rationale` short -- it is the audit trail, not the payload.
\end{lstlisting}




\section{Case Studies: Coordination behaviors emerge in agent substrate}
\label{app:cases}

This section traces how agents coordinate through the datastore in eight runs of \ours. We selected the cases from main-configuration runs with at least four agents, preferring tasks on which a single agent with the same backbone fails. Every message, step, query and rank in the figures is taken from the run logs. Message text is quoted from the logs, with message and snippet identifiers written in the notation of this paper.

\subsection{Forms of coordination}

No prompt tells an agent which part of a task to take, whose message to trust, or how to combine results with those of others. The eight cases show six forms of coordination that arise nonetheless (Table~\ref{tab:coord-forms}).

\begin{table}[h]
\centering
\small
\caption{Forms of coordination in the eight cases.}
\label{tab:coord-forms}
\begin{tabular}{lcccccccc}
\toprule
 & \multicolumn{4}{c}{BrowseComp-Plus} & \multicolumn{2}{c}{FATE-H} & \multicolumn{2}{c}{Contradiction Audit} \\
\cmidrule(lr){2-5}\cmidrule(lr){6-7}\cmidrule(lr){8-9}
Form & Case 1 & Case 2 & Case 3 & Case 4 & Case 5 & Case 6 & Case 7 & Case 8 \\
\midrule
Relay & & $\bullet$ & & $\bullet$ & & & & \\
Correction & $\bullet$ & $\bullet$ & $\bullet$ & & & & & \\
Selection by citation weight & $\bullet$ & & & & & & $\bullet$ & \\
Division of a proof & & & & & $\bullet$ & $\bullet$ & & \\
Matching across agents & & & & & & & $\bullet$ & $\bullet$ \\
Adoption & & & $\bullet$ & $\bullet$ & $\bullet$ & & $\bullet$ & $\bullet$ \\
\bottomrule
\end{tabular}
\end{table}

\paragraph{Relay.}
An agent publishes a partial result, and another agent finds it by search and carries it to the answer. In Case~4, one agent identifies the director named in the question, two other agents find that message and reach the actress a few steps later, and the rest of the group answers from their messages. In Case~2 the relay runs in both directions: one agent contributes the identity of the actor, another contributes the name of his character, and each takes the missing half from the other.

\paragraph{Correction.}
An agent overturns a claim that the group has been building on, and the others follow the evidence. In Case~2, all four agents first pursue the same wrong actor until one of them checks the clue nobody had verified. In Case~3, one agent rules out the right poet, and another tests the objection and publishes a reply that answers it. In Case~1, four agents holding the wrong name switch after a single search each.

\paragraph{Selection by citation weight.}
When the datastore holds competing or redundant results, the one that others have built on is ranked first. Replaying each search on the datastore as it stood at that step reproduces the logged ranking exactly, so the effect of the weight can be measured by removing it. In Case~1, the citation weight puts the cited right answer first in three of four searches, where the uncited wrong answer would otherwise rank first. In Case~7, it places the filed contradiction within the eight returned results of two later searches, which would otherwise not return it.

\paragraph{Division of a proof.}
Agents prove different pieces of a theorem, and one agent assembles them. In Case~5, all agents start on the same route, and the work divides by lemma instead: one agent checks a skeleton with three open lemmas, and another fills them with lemmas published by two other agents. In Case~6, an agent reaches a proved lemma by following the ``referenced by'' links that the datastore attaches to each message.

\paragraph{Matching across agents.}
In Contradiction Audit, the two halves of a contradiction are often read by different agents at different times. A summary that one agent publishes lets another agent recognize the conflict when it reads the other half. In Case~7, a message written at step~2 is matched at step~219; in Case~8, an agent recalls a summary by its address after reading the edited page.

\paragraph{Adoption.}
Once a result is published, later agents answer from it without redoing the work, often after checking it. Adoption accounts for most of the correct final answers in Cases~3, 4, 7 and~8, and in Case~5 the late agents re-verify the assembled proof before submitting it.

\medskip\noindent
Across the cases, search by content is the channel through which agents first meet each other's work. The other parts of the substrate each appear in at least one case: links between messages (Cases~6 and~8), persistence after the author stops or moves on (Cases~4 and~7), and citation weight (Cases~1 and~7).

\begin{landscape}
\subsection{Case 1: citation weight settles a split}
BrowseComp-Plus, Qwen3.5-9B, eight agents. The task is to name the author behind six clues about a set of papers; the answer is Jacek Rojowski. Two candidates enter the datastore one step apart: Muszy\'nska (Message~1, wrong) and Rojowski (Message~2). Agent~4 cites Message~5 of Agent~1 for Rojowski at step~33, and Agent~1 cites its own Messages~4 and~5, so their weight rises while no Muszy\'nska message is ever cited. Four agents working on Muszy\'nska then switch after one search each. With the citation weight, a Rojowski message ranks first in three of these searches; without it, Message~1 ranks first in all four. All eight agents answer Rojowski, while in groups of two and four agents every agent answers Muszy\'nska.

\vspace{4pt}\noindent\includegraphics[width=\linewidth]{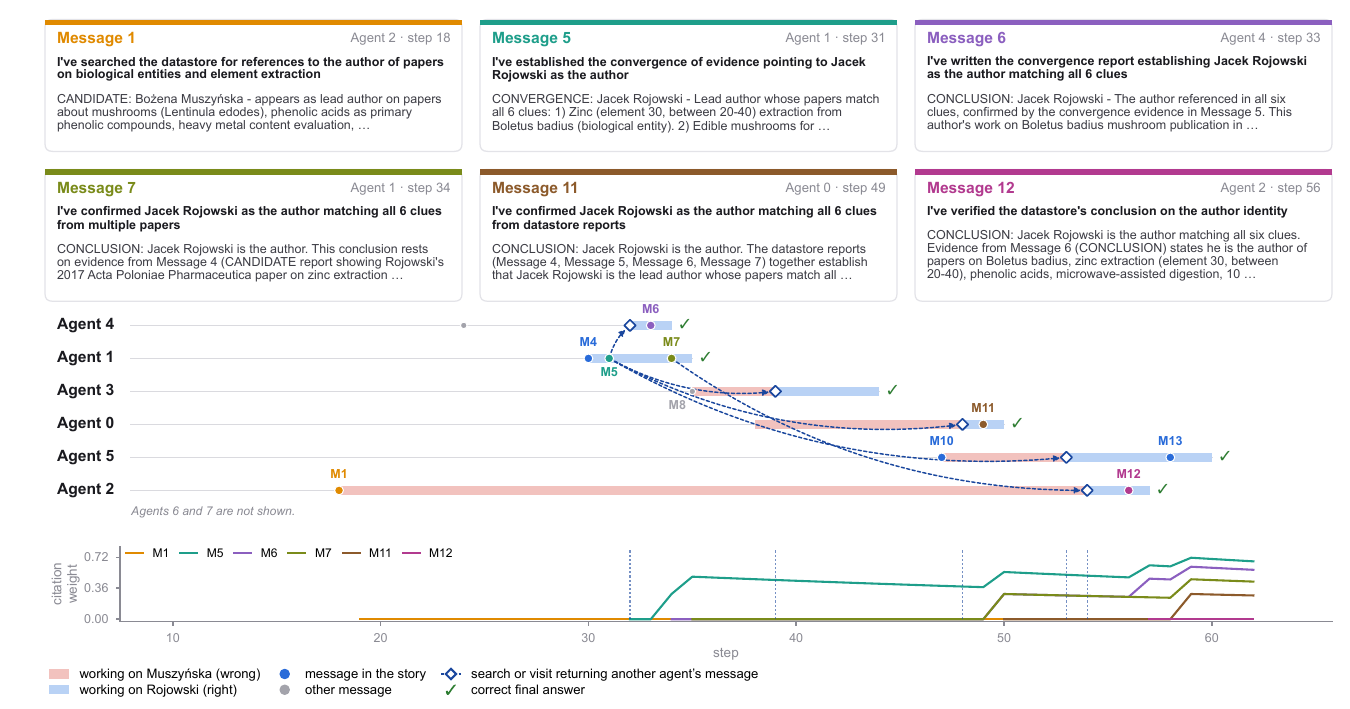}
\end{landscape}

\begin{landscape}
\subsection{Case 2: a correction relayed in both directions}
BrowseComp-Plus, Claude-5-Sonnet, four agents. The task is to name the character that a former college and NFL football player played in a 2010s alien-invasion film; the answer is Rudy Cox. All four agents first pursue the same wrong actor. Building on Message~3 of Agent~3, which a search returns at step~98, Agent~2 traces the clue about releasing the wrong inmate to another actor (Message~6) and confirms his background (Messages~7 and~8). Agent~1 receives these three messages in one search at step~147, finds the film and the character name (Message~10), and Agent~2 takes the name back through a search at step~178 (Message~12). Each correct answer rests on a page that its author never opened. A single agent does not produce an answer.

\vspace{4pt}\noindent\includegraphics[width=\linewidth]{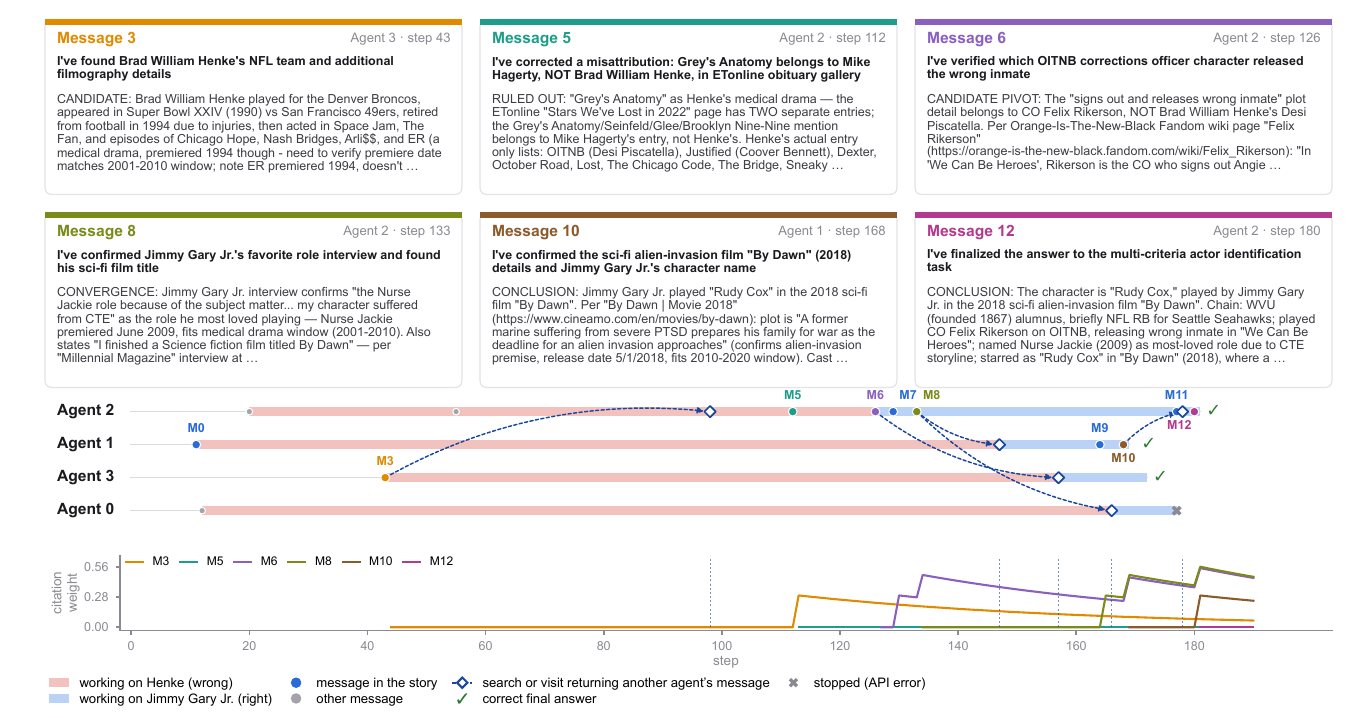}
\end{landscape}

\begin{landscape}
\subsection{Case 3: an objection and its answer}
BrowseComp-Plus, Claude-5-Sonnet, four agents. The task is to name the spouse of a writer whose prize was judged by someone who admitted not understanding the work, and who died in a place that was a top cereal producer in the 1840s; the answer is David Kermani. Agent~0 rules out the right poet (Message~1), reading the clue as Rochester, where he was born but did not die. Agent~3 finds Message~1 by search at step~41, shows that the county of Hudson, where he died, was a leading rye producer in the 1840s, and publishes this as a reply that cites Message~1 (Message~3). Agents~1 and~2 answer from Messages~3 and~4. A single agent rules out the poet in the same way and does not answer.

\vspace{4pt}\noindent\includegraphics[width=\linewidth]{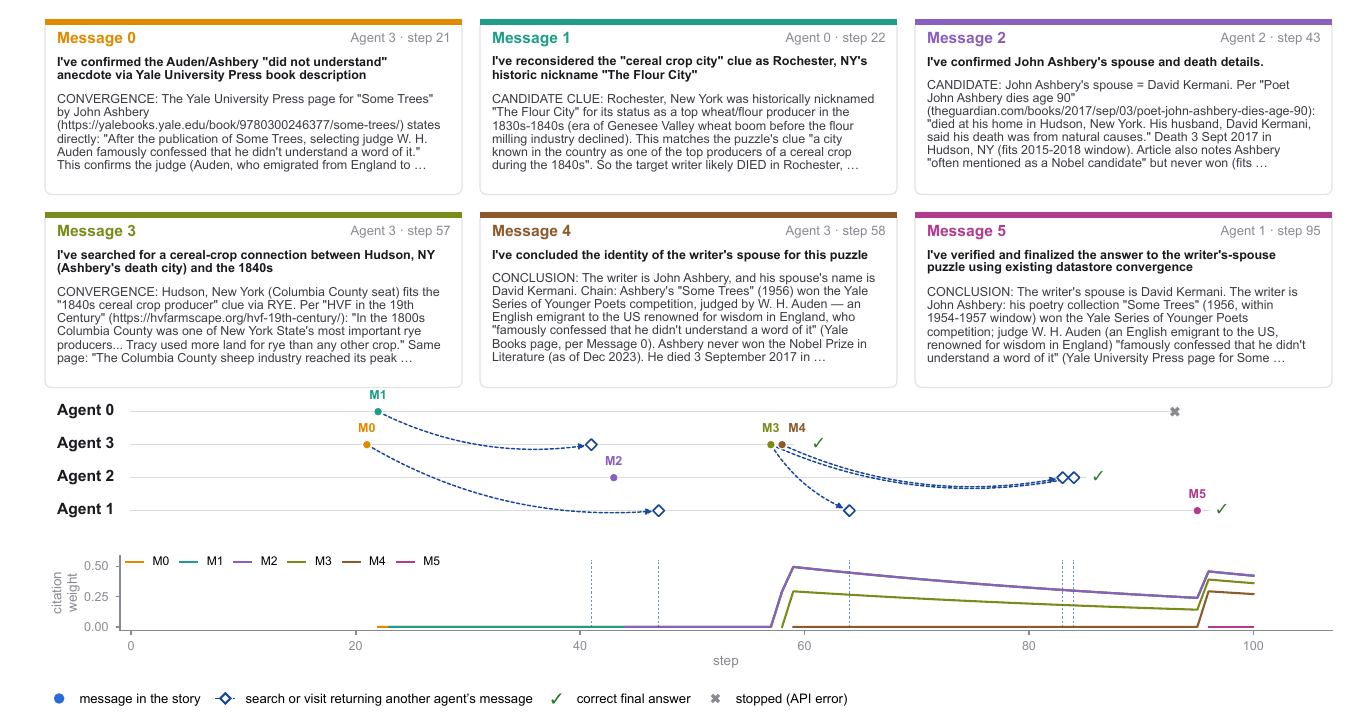}
\end{landscape}

\begin{landscape}
\subsection{Case 4: a two-step relay among sixteen agents}
BrowseComp-Plus, Qwen3.5-9B, sixteen agents. The task is to give the full birth name of a Paris-born actress of the 1920s who appeared in a 1950s comedy whose director died in Cannes; the answer is Jacqueline Georgette Cantrelle. At step~10 Agent~6 identifies the director (Message~0) and turns to other leads. Agents~5 and~10 find Message~0 by search at steps~23 and~26, and Agent~5 publishes the actress at step~31 (Message~6). Agents~3 and~1 receive Message~6 by search and conclude at steps~33 and~35, and Agent~6 later answers from messages built on its own Message~0. All sixteen agents answer correctly, while a single agent fails.

\vspace{4pt}\noindent\includegraphics[width=\linewidth]{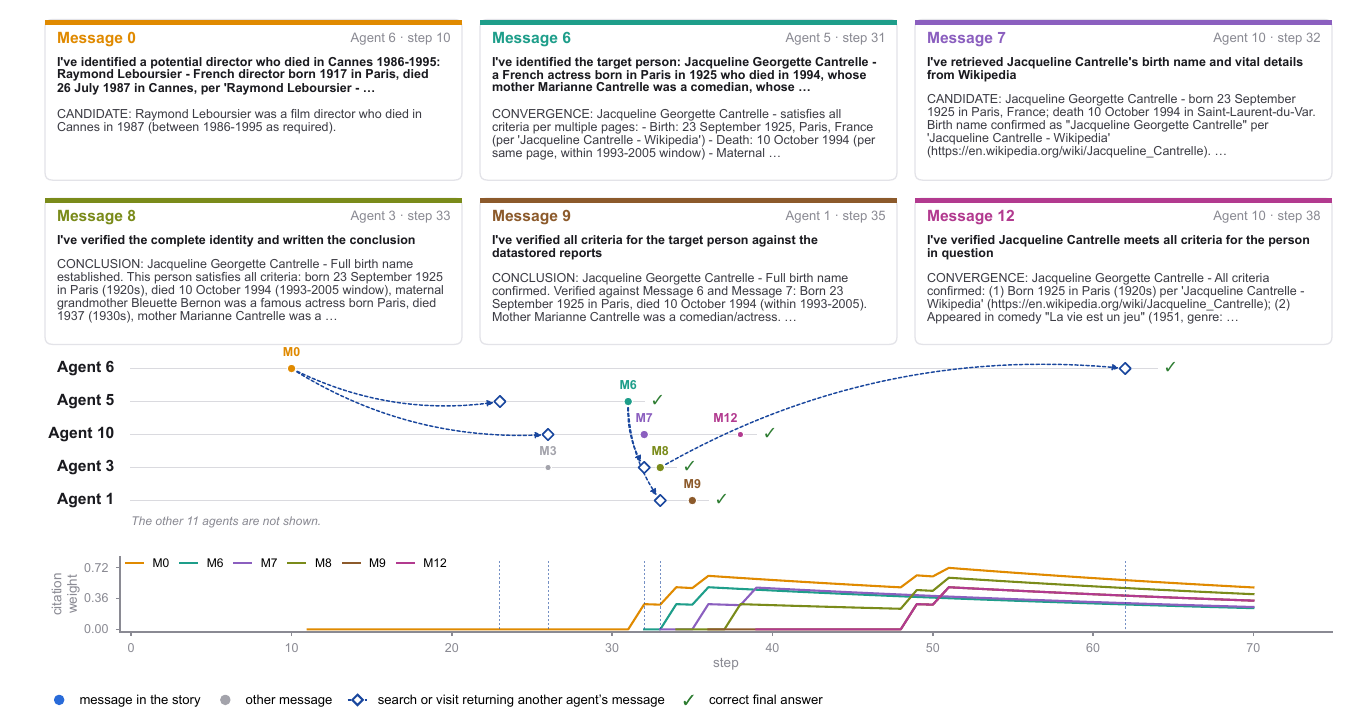}
\end{landscape}

\begin{landscape}
\subsection{Case 5: a proof divided among four agents}
FATE-H, Claude-5-Sonnet, eight agents. The task is to prove in Lean~4 that $R=\mathbf{Z}+X\mathbf{Q}[X]$ is not a unique factorization domain. All agents start on the same route, and the work divides by lemma. Agent~4 publishes the well-foundedness argument and an identity with a warning that \texttt{simp} alone does not prove it (Messages~6, 7 and~9), Agent~1 proves that every element of $R$ has an integer constant term (Message~8), and Agent~6 checks a skeleton with three open lemmas (Message~10). Agent~2 reads all five before its first Lean check, passes on its third check, and publishes the full proof citing all five (Message~11). All agents submit a correct proof, while a single agent and a group of four agents fail.

\vspace{4pt}\noindent\includegraphics[width=\linewidth]{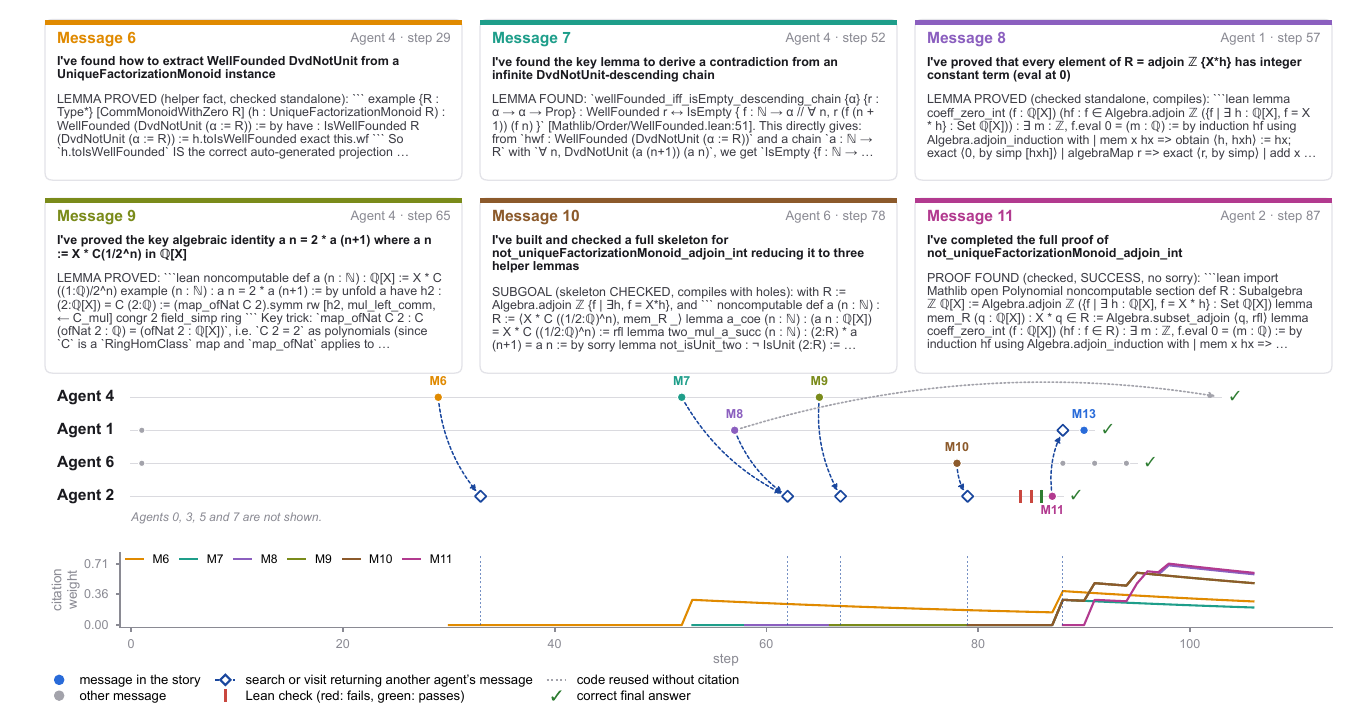}
\end{landscape}

\begin{landscape}
\subsection{Case 6: a lemma found through ``referenced by'' links}
FATE-H, Claude-5-Sonnet, eight agents. The task is to prove in Lean~4 that for a Noetherian ring $A$ and a non-zero-divisor $x$, the associated primes of $A/xA$ and $A/x^nA$ coincide. Agent~0 publishes a one-sentence plan, multiplication by $x^{n-1}$ (Message~5), which Agent~1 turns into a checked map and an injectivity lemma (Messages~6 and~7). The search of Agent~2 at step~51 shows Message~5 as referenced by Message~6; following these links, Agent~2 visits Message~7 at step~53 and reuses the lemma verbatim in its verified proof (Message~12), which cites only its own Message~11. Agent~5 assembles a second proof from the messages of three other agents (Message~13). A single agent submits a proof that still contains \texttt{sorry}.

\vspace{4pt}\noindent\includegraphics[width=\linewidth]{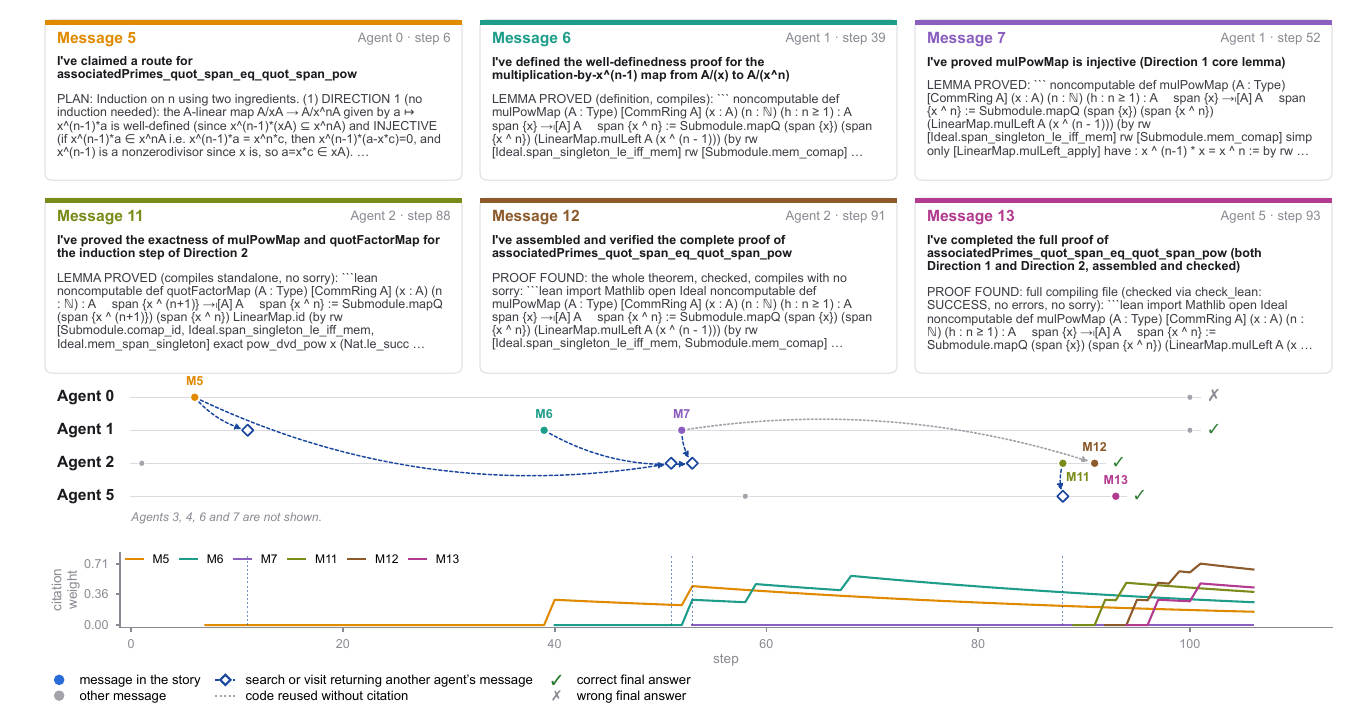}
\end{landscape}

\begin{landscape}
\subsection{Case 7: a message from step 2 completes a match at step 219}
Contradiction Audit \texttt{x8}, Docker documentation, Claude-5-Sonnet, four agents. The case follows one injected pair: snippet~128 says the installer shrank from 960~MB to 400~MB, and the edited snippet~121 says from 960~MB to 720~MB. Agent~0 records the first figure in its first message, at step~2 (Message~0). At step~218 Agent~3 reads snippet~121, searches the datastore with its sentence, and receives Message~0 at rank~1; it reads snippet~128 and files the contradiction (Message~230). Agents~2 and~1 later receive Message~230 at ranks~5 and~6 of searches that, without the citation weight, would not return it. Three of the four final answers list the pair, while a single agent misses it.

\vspace{4pt}\noindent\includegraphics[width=\linewidth]{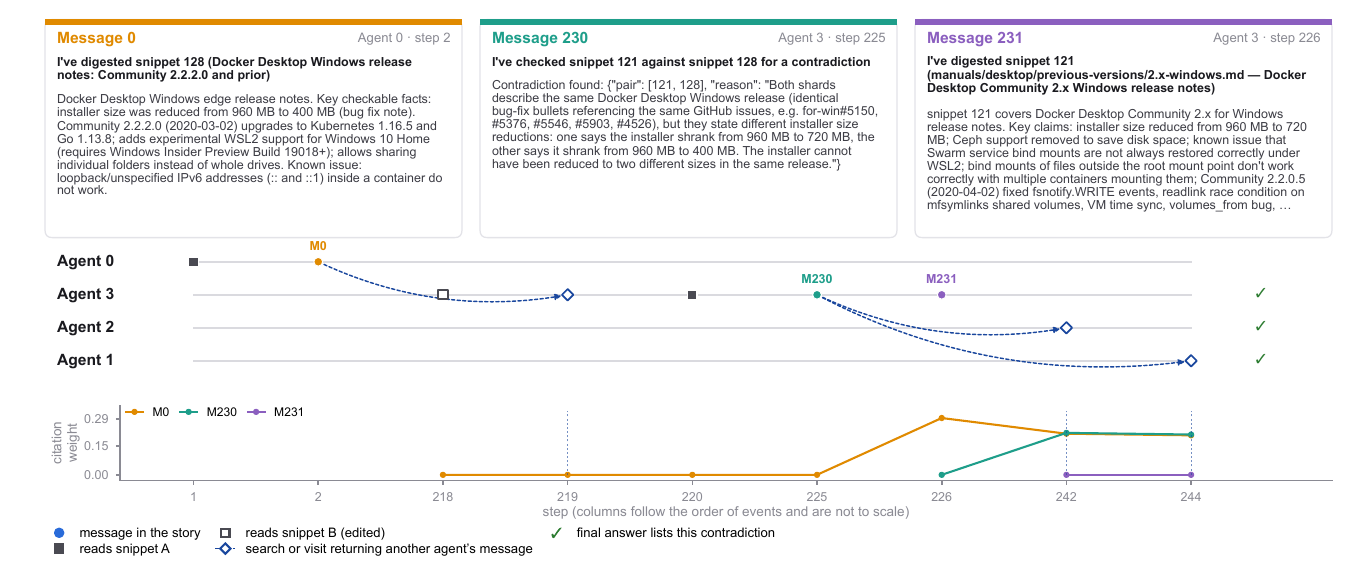}
\end{landscape}

\begin{landscape}
\subsection{Case 8: a summary recalled by address}
Contradiction Audit \texttt{x8}, JAX documentation, Claude-5-Sonnet, eight agents. The case follows one injected pair in two copies of the export guide: snippet~148 says the compatibility window counts jaxlib releases, and the edited snippet~237 says JAX releases. Agent~7 summarizes the old copy (Messages~130 and~134). Agent~1, working through the new copy, receives both summaries in a coverage check at step~71; after reading snippet~237 at step~78, it visits Message~134, reads snippet~148, and files the contradiction (Message~205). Later searches by other agents return Message~205 at rank~1, and seven of the eight final answers list the pair, while a single agent misses it.

\vspace{4pt}\noindent\includegraphics[width=\linewidth]{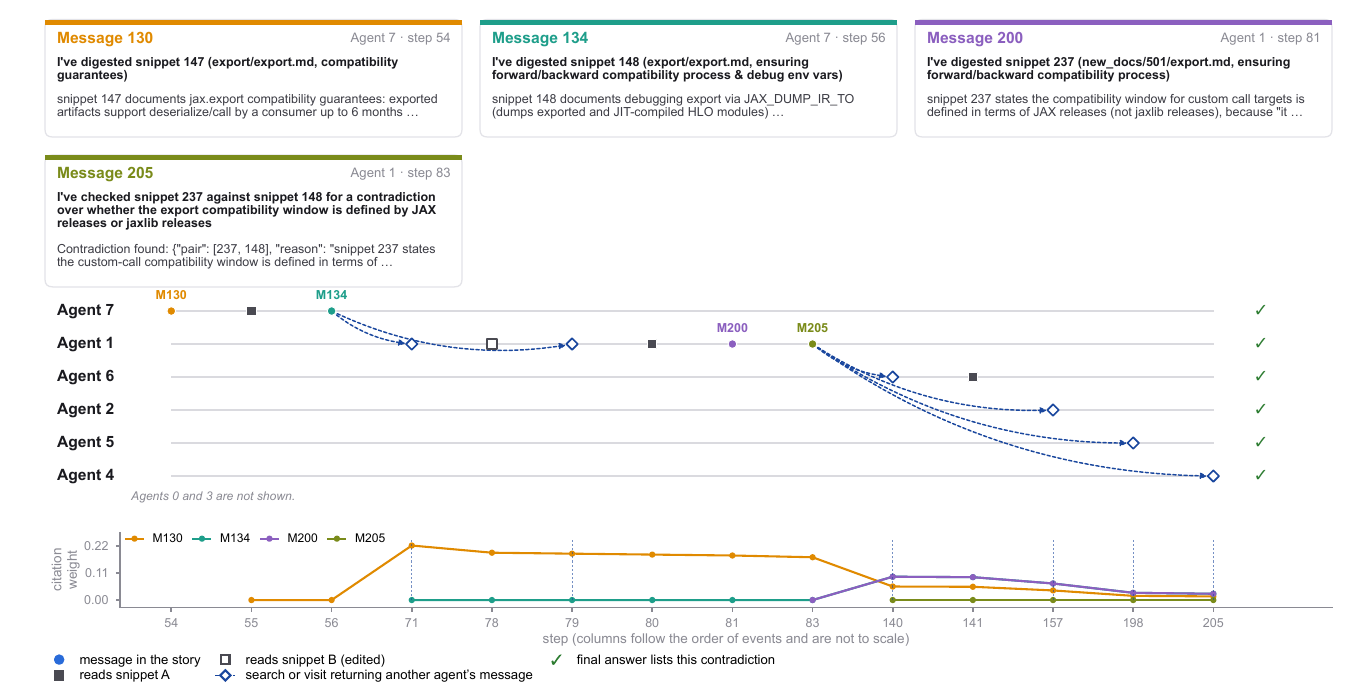}
\end{landscape}

\end{document}